\documentclass[letterpaper,journal]{IEEEtran}
\usepackage{amsmath,amsfonts}
\usepackage{algorithmic}
\usepackage{algorithm}
\usepackage{array}
\usepackage[caption=false,font=normalsize,labelfont=sf,textfont=sf]{subfig}
\usepackage{textcomp}
\usepackage{stfloats}
\usepackage{url}
\usepackage{verbatim}
\usepackage{graphicx}
\usepackage{cite}
\usepackage{hyperref}

\usepackage{amsthm}
\usepackage{fdsymbol}
\usepackage{tabularx}
\usepackage{multirow}
\usepackage{booktabs}

\theoremstyle{plain}
\usepackage{newfloat}
\usepackage{listings}

\usepackage{microtype}      

\usepackage[normalem]{ulem}

\newtheorem{definition}{Definition}

\newtheorem{theorem}{Theorem}

\newtheorem{assumption}{Assumption}

\usepackage{xurl}

\usepackage{float}  
\usepackage{multirow} 
\usepackage{makecell}
\usepackage{xcolor}
\usepackage{enumitem}
\usepackage{listings}
\usepackage{bbm}
\usepackage{bm}

\renewcommand{\nabla}{\triangledown}

\def\Vbar{{\perp\!\!\!\perp}}

\def\bE{{\mathbb E}}

\begin{document}

\title{Causal Effect Estimation under Networked Interference without Networked Unconfoundedness Assumption}

\author{Weilin Chen, 
        Ruichu~Cai\textsuperscript{*},~\IEEEmembership{Senior Member,~IEEE,}
        Jie Qiao,
        Yuguang Yan,
        José Miguel Hernández-Lobato
\thanks{This research was supported in part by National Science and Technology Major Project (2021ZD0111501),  Natural Science Foundation of China (U24A20233) and CCF-DiDi GAIA Collaborative Research Funds (CCF-DiDi GAIA 202618).}
\thanks{Weilin Chen is with the School of Computer Science, Guangdong University of Technology, Guangzhou, 510006, China (e-mail: chenweilin.chn@gmail.com).}
\thanks{Ruichu Cai is with the School of Computer Science, Guangdong University of Technology, Guangzhou, 510006, China and Peng Cheng Laboratory, Shenzhen, China (e-mail: cairuichu@gmail.com).}
\thanks{Jie Qiao and Yuguang Yan are with the School of Computer Science, Guangdong University of Technology, Guangzhou, 510006, China (e-mail: qiaojie.chn@gmail.com and ygyan@gdut.edu.cn).}
\thanks{José Miguel Hernández-Lobato is with the Department of Engineering,
University of Cambridge, Cambridge CB2 1PZ, United Kingdom (e-mail: jmh233@cam.ac.uk).}
\thanks{Corresponding author: Ruichu Cai.} 
\thanks{Manuscript received XXX; revised XXX.}
}


\markboth{Journal of \LaTeX\ Class Files,~Vol.~14, No.~8, August~2021}%
{Shell \MakeLowercase{\textit{et al.}}: A Sample Article Using IEEEtran.cls for IEEE Journals}


\maketitle

\begin{abstract}

Estimating causal effects under networked interference from observational data is a crucial yet challenging problem.
Most existing methods mainly rely on the networked unconfoundedness assumption, which guarantees the identification of networked effects.
However, this assumption is often violated due to the latent confounders inherent in observational data, thereby hindering the identification of networked effects.
To address this issue, we leverage the rich interaction patterns between units in networks, which provide valuable information for recovering these latent confounders.
Building on this insight, we develop a confounder recovery framework that explicitly characterizes three categories of latent confounders in networked settings: those affecting only the unit, those affecting only the unit's neighbors, and those influencing both.
Based on this framework, we design a networked effect estimator using identifiable representation learning techniques.
From a theoretical standpoint, we prove the identifiability of all three types of latent confounders and, by leveraging the recovered confounders, establish a formal identification result for networked effects.
Extensive experiments validate our theoretical findings and demonstrate the effectiveness of the proposed method.

\end{abstract}


\section{Introduction}
\label{intro}

\begin{figure}[!t]
    \centering
    \includegraphics[width=0.5\textwidth]{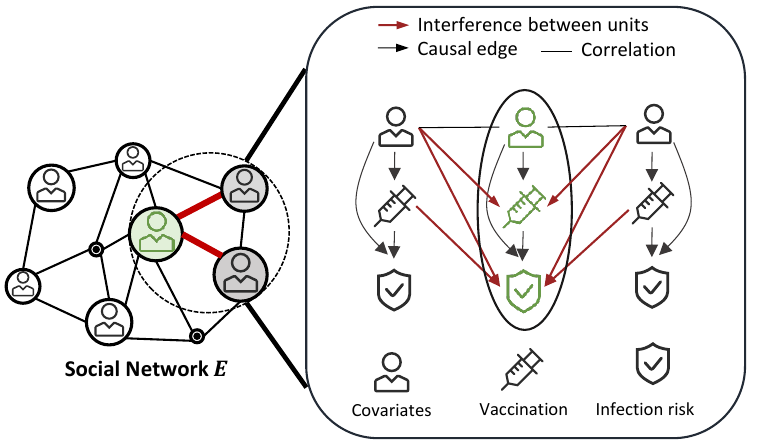}
    \caption{
    A simple example illustrating networked interference among three units.
    The networked interference introduces interactions between units, represented by the solid red arrows.
    These interactions violate the traditional SUTVA assumption, rendering the networked effects non-identifiable.
    } 
    \label{fig: example of dr} \label{fig: intro example}
\end{figure}

\IEEEPARstart{E}{stimating} causal effects in the presence of networked interference is a crucial yet challenging problem across a wide range of domains, including human ecology \cite{ferraro2019causal}, advertising \cite{parshakov2020spillover}, and epidemiology \cite{barkley2020causal}.
The core difficulty arises because interference among units inherently violates the Stable Unit Treatment Value Assumption (SUTVA), a foundational assumption of classical causal inference.
Consider, for example, evaluating the effect of a flu vaccine on individual infection rates. A standard causal inference approach assumes that an individual's infection risk depends only on their own vaccination status. 
However, in reality, vaccination can induce herd immunity: vaccinating one person may reduce the infection risks of others, leading to indirect or spillover effects (Figure \ref{fig: intro example}).
Such a violation of SUTVA introduces bias into traditional causal inference methods, rendering classical estimands inapplicable \cite{forastiere2021identification} to the networked setting.
To address this, previous causal inference studies \cite{forastiere2021identification,ma2021causal, cai2023generalization, chen2024doubly} model interference explicitly and focus on three categories of networked effects: \textit{main effects} (the impact of a unit's own treatment on itself), \textit{spillover effects} (the impact of a unit's treatment on other units), and \textit{total effects} (the combined impact of both).

Under networked interference, to identify and estimate the above three estimands, existing methods rely on a key assumption named \textit{networked unconfoundedness}.
This assumption states that all relevant confounders are captured by the observed covariates and those of neighboring units, with no additional latent factors influencing both treatment and outcome.
Based on this assumption, various methods have been developed for networked effect identification and estimation, including generalized propensity score methods \cite{forastiere2021identification}, balanced representation methods \cite{chin2019regression, ma2021causal, cai2023generalization}, and doubly robust methods \cite{liu2019doubly, chen2024doubly}.

\begin{figure}[!t]
    \centering
    \includegraphics[width=0.48\textwidth]{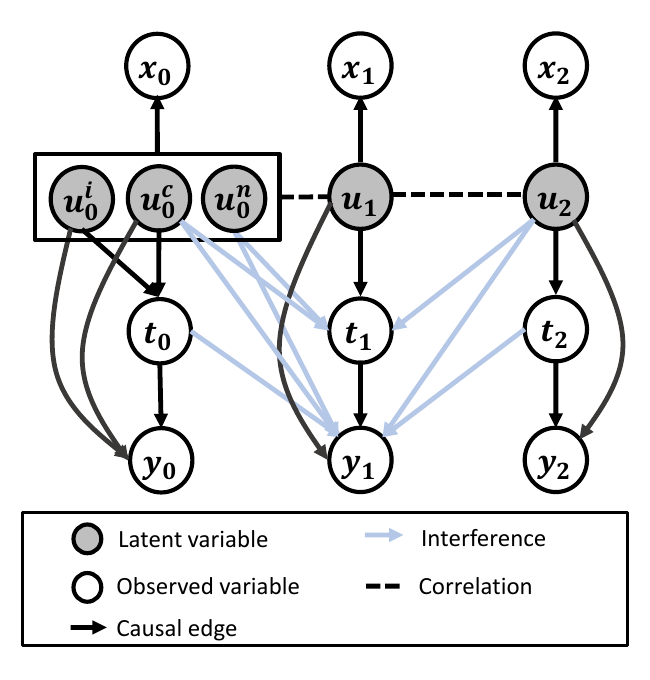}
    \caption{Assumed causal graph in this paper.
    $x$ denotes observed proxies, $u$ denotes latent confounders, $t$ denotes the treatment, and $y$ denotes the outcome of interest.
    We assume that latent confounders $u$ contain three types of variables, i.e., $u^i$ affecting the unit itself, $u^n$ affecting the unit's neighbors, and $u^c$ affecting both.} 
    \label{fig: causal graph} 
\end{figure} 

However, the networked unconfoundedness assumption is frequently violated in real-world scenarios relying on observational data, thereby significantly limiting the applicability and reliability of existing methods.
For example, in the case of flu vaccination, an individual’s decision to vaccinate may be influenced by socioeconomic factors such as personal income or household financial status.
These socioeconomic factors, however, are often difficult to directly measure due to privacy restrictions or data availability constraints.
The presence of such latent confounders, violating the networked unconfoundedness assumption, biases both main and spillover effect estimation, as well as total effect estimation, thereby fundamentally limiting existing approaches.

To address this limitation, we propose a framework that eliminates the need for the networked unconfoundedness assumption.
Specifically, we begin by exploring three categories of latent confounders, illustrated in Figure \ref{fig: causal graph}, that hinder effect identification: $u^i$ affecting only the unit, $u^n$ affecting only neighbors, and $u^c$ influencing both.
Rather than assuming networked unconfoundedness, we investigate the identifiability of latent confounders in the presence of networked interference.
A key insight of our work is that networked interference itself provides auxiliary information that can be exploited to recover latent confounders.
Building on this recovery, we theoretically establish the networked effect identification result and further devise a practical effect estimator under networked interference.
Overall, our contribution can be summarized as follows:
\begin{itemize}
    \item We address the problem of networked effect identification and estimation in the presence of latent confounders, and categorize three types of latent confounders that hinder identification.
    \item We establish the identifiability of latent confounders by leveraging the auxiliary information revealed by networked interference, and prove that networked effects can be identified.
    \item We devise a networked effect estimator built upon these theoretical findings. Extensive experiments validate our theoretical results and demonstrate the effectiveness of the proposed method.
\end{itemize}

\section{Related Works}
\label{related works}
\textbf{Traditional Causal Inference} has been widely used in several scientific domains, e.g., \cite{gangl2010causal, glass2013causal, Wu2022OnTO, huang2025learning}, as it can directly measure the effect of an intervention on the outcome \cite{rubin1974estimating, pearl2009causality}.
Classic methods usually assume that all confounders can be observed and units do not affect each other, which are called the unconfoundedness assumption and SUTVA, respectively. 
Under these assumptions, two kinds of known methods are the inverse probability weighting estimators (IPW) \cite{IPWrosenbaum1983central, IPWrosenbaum1987model} and the outcome regression estimators \cite{kunzel2019metalearners}. 
These methods have many variants and applications, e.g. \cite{rosenbaum1985constructing, li2018balancing, CAI2024106336, assaad2021counterfactual, johansson2016learning, 9133279, 9712445}.
Combining the two, the doubly robust estimators are also well studied \cite{robins1994estimation}, which are robust to model misspecification.
\underline{Different from traditional causal inference estimators,} we focus on causal effect identification and estimation, without the unconfoundedness assumption and SUTVA.

\textbf{Causal Inference without SUTVA} has received increasing attention in recent years, broadening the applicability of traditional causal inference methods.
Many existing methods extend classical causal inference to networked settings.
The unconfoundedness assumption is extended to the networked unconfoundedness assumption by additionally including the treatment and covariates of neighboring units.
Based on the extended assumptions, \cite{liu2016inverse} proposes several weighted estimators based on a generalized IPW under interference.
\cite{forastiere2021identification} introduces the generalized propensity score for networked effect identification and estimation.
Further assuming the network can be decomposed into several components, \cite{lee2021estimating} proposes a closed-form estimator for the asymptotic variance.
To achieve a more accurate effect estimation, several works consider balancing representation learning.
Prior works achieve this through different strategies, including incorporating neighborhood features with adversarial learning \cite{jiang2022estimating}, Hilbert-Schmidt independence criterion (HSIC) score \cite{ma2021causal}, and reweighted representation learning \cite{cai2023generalization}.
Additionally,\cite{mcnealis2023doubly, liu2023nonparametric, chen2024doubly} propose networked effect estimators to achieve the double robust (DR) property.
Moreover, \cite{zhao2024learning} proposes using an attention mechanism to capture heterogeneous interference. 
\underline{Different from them}, we explore the problem of networked effect estimation without the networked unconfoundedness assumption.

\textbf{Causal Inference without Unconfoundedness Assumption} is an important problem since the unconfoundedness assumption is usually violated in observational studies.
Classic methods to solve this problem usually assume there exist additional variables, e.g., instrumental variables \cite{pearl2000models,stock2003retrospectives,wu2022instrumental,cheng2024instrumental,cheng2024disentangled}, proximal variables \cite{miao2018identifying, tchetgen2024introduction}.
Another effective way is to learn substitutes for the confounders or directly recover the latent confounder using representation learning methods.
For example, \cite{wang2019blessings} and its variants \cite{bica2020time, huang2025multi} propose using multiple causes to learn factor models as valid substitutes for confounders.
CEVAE \cite{louizos2017causal} assumes that latent confounders can be recovered by their proxies and applies VAE to learn confounders.
As a follow-up work, TEDVAE \cite{zhang2021treatment} and DMAVAE \cite{xu2023disentangled}
decouple the learned latent confounders into several factors for treatment effect estimation in different settings.
Our work is closely related to these works.
Most of these methods are developed for conventional non-networked settings. In networked settings, the corresponding networked unconfoundedness assumption can also be violated due to latent confounding involving both the unit and its neighbors. SPNet \cite{huang2023modeling} learns representations of latent confounders, 
while DisIV \cite{huang2026disentangled} recovers latent instrumental variables through structural disentanglement.
\underline{Different from them}, we directly recover role-specific latent confounders for networked effect estimation and establish their identifiability, providing theoretical guarantees for the proposed estimator.

\section{Notations, Assumptions, Estimands}
\label{notations, assumptions, estimands}

In this section, we first introduce the notations used in this work. 
Our notations follow the potential outcome framework \cite{rubin1974estimating}.
Let $U \in \mathcal{U}$ denote the set of latent confounders,
and also let $X \in \mathcal{X}$ represent observable proxies of latent confounders.
Let $T \in \{0,1\}$ be a binary treatment, where $T=1$ indicates a unit receives the treatment (treated) and $T=0$ indicates that the unit remains untreated (control). 
Let $Y \in \mathcal{Y}$ denote the real-valued outcome.

Throughout, we use lowercase letters (e.g., $x,u,y,t$) to denote the value of random variables,
and subscripted lowercase letters (e.g., $x_i,u_i,y_i,t_i$) denote the realization corresponding to the $i$-th unit.
Then, the networked dataset is denoted as $D=(\{x_i,t_i,y_i\}_{i=1}^n,E)$, where $E$ is the adjacency matrix of the network and $n$ is the total number of units. 
The set of first-order neighbors of unit $i$ is denoted by $\mathcal{N}_i$, with their treatment, confounders, and covariate vectors denoted by $t_{\mathcal N _i}$, $u_{\mathcal N _i}$, and $x_{\mathcal N _i}$, respectively.
Given the network structure, we decompose the latent confounders of unit $i$ as $u_i=(u_i^i,u_i^n,u_i^c)$, where $u_i^i$ affects only the unit $i$ itself, $u_i^n$ affects only its neighbors, and $u_i^c$ affects both the unit and its neighbors, as illustrated in Figure \ref{fig: causal graph}.

In the presence of networked interference, a unit’s potential outcome depends not only on its own treatment assignment but also on the treatment assignments of its neighbors.
Formally, following \cite{forastiere2021identification}, we denote the potential outcome of unit $i$ as $y_i(t_i, t_{\mathcal{N}_i})$, and then the realized outcome $y_i$ corresponds to the factual outcome.
Following \cite{forastiere2021identification, cai2023generalization, chen2024doubly}, we further assume that the dependence of the potential outcome on the neighbors’ treatments operates through a specified summary function $g$: $\{0,1\}^{|\mathcal{N}_i|}\rightarrow [0,1]$.
Let $z_i$ denote the neighborhood exposure derived from this function. 
Note that $g$ can, in principle, be learned from data; however, learning $g$ is beyond the scope of this work.
In particular, we aggregate the treatment assignments of neighbors using the average exposure, i.e., $z_i=\frac{\sum_{j\in \mathcal{N}_i}t_j}{|\mathcal{N}_i|}$. 
Under this specification, the potential outcome $y_i(t_i,t_{\mathcal N _i}) $ can be equivalently expressed as $y_i(t_i,z_i)$,
indicating that, under networked interference, each unit is subject to two distinct forms of treatment: the binary individual treatment $t_i$ and the continuous neighborhood exposure $z_i$.

In this paper, the \textbf{goal} is to estimate the average dose-response function and the conditional average dose-response function, which are formally defined as 
\begin{equation}
  \begin{aligned}
    & \psi(t,z) := \mathbb E [Y(t,z)], \\
    & \mu(t,z,x,x_{\mathcal N }) := \mathbb E [Y(t,z)|X=x,X_{\mathcal N }=x_{\mathcal N }],
    \end{aligned} 
\end{equation}
where the expectation in $\psi(t,z)$ is taken over the target population of units. For a finite population, it corresponds to averaging potential outcomes over all units, whereas $ \mu(t,z,x,x_{\mathcal N })$ conditions on the observed proxies of the focal unit and its neighbors.

Based on the definitions of (conditional) average dose-response function, existing works \cite{forastiere2021identification, ma2021causal, cai2023generalization, chen2024doubly} primarily focus on the following different kinds of causal effects:

\begin{definition}[Average Effects]
Average effects quantify changes in the expected outcomes due to interventions on treatment and neighborhood exposure:
\begin{itemize}
    \item \textbf{Average Main Effect (AME):} 
    $\tau^{(t,0),(t',0)} = \psi(t,0) - \psi(t',0)$, the effect of changing the individual treatment while holding neighborhood exposure fixed.
    \item \textbf{Average Spillover Effect (ASE):} 
    $\tau^{(0,z),(0,z')} = \psi(0,z) - \psi(0,z')$, the effect of changing neighborhood exposure while holding the individual treatment fixed.
    \item \textbf{Average Total Effect (ATE):} 
    $\tau^{(t,z),(t',z')} = \psi(t,z) - \psi(t',z')$, the combined effect of changing both individual treatment and neighborhood exposure.
\end{itemize}
\end{definition}

\begin{definition}[Individual Effects]
Individual effects quantify changes in the conditional expected outcomes of a specific unit $x_i$ given its and neighbors' covariates:
\begin{itemize}
    \item \textbf{Individual Main Effect (IME):} 
    $\tau_i^{(t,0),(t',0)} = \mu(t,0, x_i, x_{\mathcal N_i}) - \mu(t',0, x_i, x_{\mathcal N_i})$, the effect of changing the unit's treatment.
    \item \textbf{Individual Spillover Effect (ISE):} 
    $\tau_i^{(0,z),(0,z')} = \mu(0,z, x_i, x_{\mathcal N_i}) - \mu(0,z', x_i, x_{\mathcal N_i})$, the effect of changing neighbors' treatments.
    \item \textbf{Individual Total Effect (ITE):} 
    $\tau_i^{(t,z),(t',z')} = \mu(t,z, x_i, x_{\mathcal N_i}) - \mu(t',z', x_i, x_{\mathcal N_i})$, the combined effect of changing both individual and neighborhood treatments.
\end{itemize}
\end{definition}

Here,  we use $0$ as the reference level following existing works \cite{forastiere2021identification, ma2021causal, cai2023generalization, chen2024doubly}. The choice is not essential, and the main effect can be defined by holding the neighborhood exposure at any fixed $z$, similarly for the spillover effect.

Throughout this paper, we also assume the following assumptions hold:

\begin{assumption}[Networked Consistency] \label{asmp: consistency}
The observed outcome coincides with the potential outcome under the realized individual treatment and neighborhood exposure. Formally, if unit $i$ receives treatment $t_i$ and neighborhood exposure $z_i$, then $y_i=y_i(t_i,z_i)$.
\end{assumption}

\begin{assumption}[Networked Overlap] \label{asmp: Overlap}
    Given any unit and neighbors' confounders, any treatment pair $(t,z)$, we have $ p(t_i,z_i|  u^i_i, u_i^c, u_{\mathcal N _i}^c , u_{\mathcal N _i}^n ) > 0 $,
    where $p$ denotes the conditional probability mass or density, as appropriate.
\end{assumption}

\begin{assumption}[Neighborhood Interference] \label{asmp: Neighborhood interference}
    The potential outcome of a unit depends only on its own treatment and the treatments of its first-order neighbors, and this dependence operates exclusively through a summary function $g$.
    Formally, $\forall t_{\mathcal N _i}$,$t^{\prime}_{\mathcal N _i}$ which satisfy $g(t_{\mathcal N _i})=g(t^{\prime}_{\mathcal N _i})$, we have $y_i(t_i, t_{\mathcal N _i})=y_i(t_i, t^{\prime}_{\mathcal N _i})$.
\end{assumption}

These assumptions are standard in the networked causal inference literature (see, e.g., \cite{forastiere2021identification, cai2023generalization, ma2022learning}).
Specifically, Assumption \ref{asmp: consistency} rules out the possibility of multiple versions of treatment, ensuring that the potential outcomes of an individual are consistent across every identical treatment–exposure pair, making causal effects well-defined.
Assumption \ref{asmp: Overlap} requires that, within the network, each unit has a nonzero probability of every treatment–exposure pair, allowing for valid causal comparisons.
Assumption \ref{asmp: Neighborhood interference} essentially allows the existence of networked interference, where this interference is typically localized—operating through direct or first-order neighbors and can be summarized by a network exposure function $g$ that captures how neighboring treatments influence the focal unit’s outcome.
The average exposure used here is a convenient specification following prior work rather than a necessary form of our framework. 
It assumes that neighbors contribute symmetrically and that their treatment assignments affect the focal outcome only through this scalar summary, which may be restrictive when edge strengths or interference mechanisms are heterogeneous. 
More generally, Assumption 3 allows $g$ to be replaced by other prespecified exposure mappings, such as weighted neighborhood exposures. 
A learnable permutation-invariant mapping, e.g., a neural aggregation function, could also be considered, provided that the resulting exposure satisfies the corresponding overlap and interference assumptions. 
Jointly learning such an exposure mapping and the causal effect estimator is left for future work.

Additionally, existing methods typically rely on the following key assumption to ensure causal effect identification:

\begin{assumption}[Networked Unconfoundedness] \label{asmp: Network unconfounderness}
    The individual treatment and neighborhood exposure are conditionally independent of the potential outcome given the covariates of the unit and its neighbors. Formally, $\forall t,z$, we have $y_i(t,z) \Vbar t_i,z_i|x_i, x_{\mathcal N _i}$.
\end{assumption}

Assumption \ref{asmp: Network unconfounderness} generalizes the classical unconfoundedness assumption to networked settings and implies that no latent confounders simultaneously influence both the potential outcome $y_i$ and the treatments $(t_i, z_i)$.
Under the assumptions above, the networked effects can be identified \cite{forastiere2021identification}. However, in real-world scenarios, it is often unrealistic to assume that all confounders are observed, making Assumption \ref{asmp: Network unconfounderness} potentially too strong.

To address this limitation, we adopt a weaker assumption that incorporates latent confounders:
\begin{assumption}[Latent Networked Unconfoundedness] \label{asmp: Latent Network unconfounderness}
    The individual treatment and neighborhood exposure are conditionally independent of the potential outcome given the latent unit's and neighbors' confounders, i.e., $\forall t,z$, we have $y_i(t,z) \Vbar t_i,z_i|u^i_i, u^c_i, u^c_{\mathcal N _i},  u^n_{\mathcal N _i}$.
\end{assumption}

This assumption has two key components.
First, it relaxes the strong requirement of fully observed confounders by allowing for latent confounders, denoted as $u^i,u^c,u^n$.
Second, it explicitly decomposes these latent confounders into three categories: $u^i_i$, which only influences the unit itself; $u^n_i$, which affects the unit’s neighbors; and $u^c_i$, which simultaneously influences both the unit and its neighbors.
Accordingly, the relevant confounder set under the networked setting consists of the unit-level $u^i, u^c$ and the neighbor-level $u^c_{\mathcal N_i}, u^n_{\mathcal N_i}$.
This decomposition motivates us to separately identify each type of latent confounder, thereby enabling more accurate estimation.
In the following section, we will discuss the identifiability of these latent confounders and how it facilitates networked effect identification and estimation.

\section{Latent Confounder Identifiability Enables Networked Effect Identification} \label{sec: Identication}

In this section, our goal is to establish networked effect identification in the presence of latent confounders $u^i,u^c,u^n$.
According to Assumption \ref{asmp: Latent Network unconfounderness}, such identification requires access to the latent confounders.  This motivates us to employ identifiable causal representation learning techniques for their recovery.
To this end, we first introduce the latent confounder identifiability (Theorem \ref{theorem:recover latent}) and further show that the recovered latent confounders can be disentangled (Theorem \ref{theorem: disentangle}). 
Building upon these results, we ultimately establish the identification of networked effects (Theorem \ref{theorem: identify}).

To begin with, following existing identifiable representation learning methods \cite{khemakhem2020variational,lu2022invariant}, we first introduce the generative model as follows:
 \begin{equation} \label{eq: generative model}
      \begin{aligned}
    p_{\bm \theta}(X,U|X_\mathcal{N}) &= p_{\bm f}(X|U)p_{\bm{T},\bm{\lambda}}(U|X_\mathcal{N}), \\ 
      p_{\bm f}(X|U) & = p_{\bm \epsilon} (X- \bm f(U)),
      \end{aligned}
\end{equation}
where $p_{\bm \epsilon}(X- \bm f(U)) = p(\bm \epsilon)$ is the distribution of additive noise $X= \bm f(U)+\bm \epsilon$. Further, we assume $p_{\bm{T},\bm{\lambda}}(U|X_\mathcal{N})$ follows the exponential family distribution as follows.

\begin{assumption} \label{asmp: exp dist}
    The correlation between  $U$ and  $X_\mathcal{N}$ is characterized by:
    \begin{equation} \label{eq: distribution model}
      \begin{aligned}
        p_{\bm{T},\bm{\lambda}}(U|X_\mathcal{N}) 
        & = \frac{\mathcal Q (U)}{\mathcal C (X_\mathcal{N})} \exp \left[ \bm{T}(U)^T \bm{\lambda}(X_\mathcal{N}) \right]
      \end{aligned}
    \end{equation}
    where $\mathcal Q$ is the base measure and $\mathcal C$ is the normalizing constant. The $\bm{\lambda}(X_{\mathcal{N}})$ is an arbitrary function, and the sufficient statistics $\bm{T}(U) = [\bm{T}_f(U)^T,\bm{T}_{MLP}(U)^T]^T$ contains a) the sufficient statistics $\bm{T}_f(U)^T=[\bm{T}_1(U_{(1)})^T, \dots, \bm{T}_{d_U}(U_{(d_U)})^T]$ of a factorized exponential family, where all the 
    $\bm{T}_i(U_{(i)})$ have dimension larger than or equal to $2$ and $d_U$ is the dimension of $U$, and b) the output $\bm{T}_{MLP}(U)$ of a neural network with ReLU activations.
\end{assumption}

\textbf{Interpretation of the generative model.} 
Eq. \eqref{eq: generative model} separates the model into an observation model and a latent prior. 
Specifically, $p_{\bm f}(X|U)$ describes how the latent confounders generate the observed proxies, where $\bm f$ maps $U$ to the noiseless component of $X$ and $\bm \epsilon$ denotes additive noise.
And $p_{\bm{T},\bm{\lambda}}(U|X_\mathcal{N})$ characterizes how the neighboring covariates $X_\mathcal{N}$ provide auxiliary information about the latent confounders.
Here, as illustrated in Eq. \eqref{eq: distribution model}, $\bm{T}(U)$ denotes the sufficient statistics of $U$ and $\bm{\lambda}(X_{\mathcal{N}})$ maps the neighboring information to the corresponding natural parameters.
Therefore, the first component links latent confounders to their observed proxies, whereas the second component introduces the auxiliary information required for identifiability.

Assumption \ref{asmp: exp dist} is introduced by \cite{lu2022invariant}. The distribution in Assumption \ref{asmp: exp dist} is more flexible than the standard assumed distribution condition in identifiable representation learning (Eq. (7) in \cite{khemakhem2020variational}).
This assumption allows for the case that the different elements of latent confounders are not independent given the conditional set. 
The term $\bm{T}_{MLP}(U)$ does capture arbitrary dependencies between latent variables since the neural network with ReLU activation has universal approximation power.

Now, we formally state the theoretical result of the identifiability of latent confounders.

\begin{theorem}\label{theorem:recover latent}
Suppose Assumption \ref{asmp: exp dist} holds, and suppose the following conditions hold: 
(1) The set $\{X \in \mathcal O| \varphi_{\bm \epsilon}(X)=0 \}$ has measure zero where $\varphi_{\bm \epsilon}$ is the characteristic function of density $p_{\bm \epsilon}$.
(2) $\bm f$ is injective and has all second-order cross derivatives.
(3)  The sufficient statistics in $\bm T_{\bm f}$ are all
twice differentiable and each $\bm T _{\bm f}$ has at least one invertible dimension. 
(4) There exist $k+1$ distinct values $x_{\mathcal N_0}, ... , x_{\mathcal N_{ k }}$ such that the matrix
    \begin{equation*}
    \begin{aligned}
         L =  (
         \bm {\lambda}(x_{\mathcal N_1})- \bm {\lambda}(x_{\mathcal N_0}), 
         ..., 
         \bm{\lambda}(x_{\mathcal N_{ k  }})-\bm{\lambda}(x_{\mathcal N_{0}}))
    \end{aligned}
    \end{equation*}
    of size $k \times k$ is invertible where $k$ is the dimension of $\bm T$. Then we learn the true latent variable $U = \left[U^i, U^c, U^n \right]$ up to a permutation and element-wise transformations.
\end{theorem}

Proof is given in Appendix A.

\textbf{Discussion on assumptions:} Assumption \ref{asmp: exp dist} indicates our theory holds for a rich family of conditional densities \cite{Wainwright2008expfamily}. The assumption on the exponential family distribution is not strong, since many well-known distributions belong to this family, including the Gaussian, Uniform, Poisson distributions, and so on. 
Conditions (1)-(4) are mild and common assumptions in causal representation learning, e.g., \cite{khemakhem2020variational, lu2022invariant, zhu2025causal}.
In particular, Condition (3) is motivated by \cite{zhu2025causal} to ensure that the identifiability holds for the recovered latent variables beyond their sufficient statistics.
Among these, the most critical is Condition (4), which requires that the auxiliary information be sufficiently informative. 
In our networked setting, this translates to the existence of enough variability in neighbors’ covariates, 
a condition that is typically satisfied when a sufficiently rich set of covariates is collected, especially if some covariates are continuous.

Theorem~\ref{theorem:recover latent} shows that, under mild assumptions, the latent confounders can be recovered up to an invertible transformation by utilizing the auxiliary information revealed by networked interference. 
Specifically, the recovered latent variables $\hat U^i, \hat U^c, \hat U^n$ satisfy $\hat U^i, \hat U^c, \hat U^n = h(U^i, U^c, U^n)$ for some bijective element-wise transformations $h$. 
However, the exact correspondence between the recovered dimensions and the original latent variables $U^i,  U^c$ and $U^n$ remains unidentified due to an unknown permutation.
This ambiguity, nevertheless, can be resolved by exploiting the asymmetric roles that these latent variables play in the causal structure: $U^i$ only affects the unit itself, $U^n$ only affects the neighbors, and $U^c$ affects both, as shown in the following theorem.

\begin{theorem}\label{theorem: disentangle}
Suppose Theorem~\ref{theorem:recover latent} holds, and the downstream outcome model adopts a predefined role-specific feature construction, where $U^i$ is used only for the unit's own outcome, $U^n$ only for neighboring outcomes, and $U^c$ for both.
Then, by maximizing the log-likelihood of $Y$, i.e.,
$\max \log p(Y \mid U^i, U^c, U^c_\mathcal{N}, U^n_\mathcal{N},T,Z),$
the latent variables $U^i$, $U^c$, and $U^n$ are identifiable up to a permutation and element-wise transformations \textit{within each group}, respectively.
\end{theorem}
 
Proof is given in Appendix B.

Theorem \ref{theorem: disentangle} indicates that the three categories of recovered latent confounders can be further disentangled, i.e., the recovered $\hat U^i, \hat U^c, \hat U^n$ satisfying $\hat U^i=h_i(U^i), \hat U^c=h_c(U^c), \hat U^n=h_n(U^n)$ for some simple functions $h_i, h_c, h_n$.
The remaining ambiguity is restricted to element-wise transformations and permutations within each group, which is sufficient for causal effect identification.
Based on Theorem \ref{theorem: disentangle} above, we can further identify networked effects as follows:

\begin{theorem}\label{theorem: identify}
Suppose Assumptions \ref{asmp: consistency},  \ref{asmp: Overlap},  \ref{asmp: Neighborhood interference}, \ref{asmp: Latent Network unconfounderness}, and Theorems \ref{theorem:recover latent} and \ref{theorem: disentangle} hold. Then the networked effects $\mu (t,z, x, x_\mathcal{N})$ can be identified by:
\begin{equation}
\begin{aligned}
    &\mu (t,z, x, x_\mathcal{N})
    \\ &= \mathbb E [\mathbb E [Y \mid \hat U^i=\hat u^i,\hat U^c=\hat u^c,\hat U^c_{\mathcal N }=\hat u^c_{\mathcal N },\hat U^n_{\mathcal N }=\hat u^n_{\mathcal N },
    \\ & \quad \quad \quad \quad \quad
    T=t,Z=z]]
\end{aligned}
\end{equation}
where $\hat{\circ}$ denotes the recovered latent variables, 
and similarly for the average effects $\psi(t,z)$
and the outer expectation is taken with respect to the conditional distribution $p(\hat U^i,\hat U^c,\hat U^c_{\mathcal N },\hat U^n_{\mathcal N } | X=x, X_{\mathcal N }=x_{\mathcal N })$.
\end{theorem}

Proof is given in Appendix C.

Theorem \ref{theorem: identify} indicates that if we can identify latent confounders, the networked effects are thereby identifiable.
This necessitates the utilization of identifiable representation learning techniques for causal inference in the presence of networked interference and latent confounders. Built upon the theoretical findings, in the next section, we propose our practical networked effect estimator that utilizes identifiable causal representation learning techniques.

\begin{figure*}[!t]
    \centering
    \includegraphics[width=1.\textwidth]{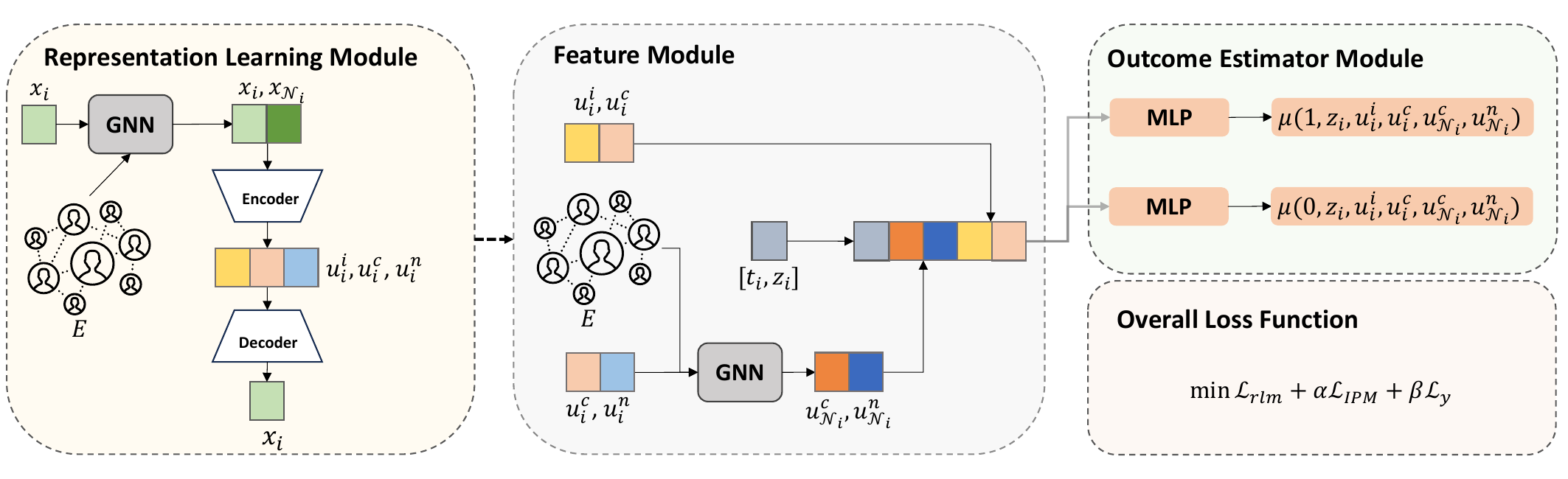}
    \caption{Model architecture of the proposed CaLaNet. The representation learning module learns the latent confounders, the feature module aggregates the confounders of unit $i$ and its neighbors, and the outcome estimator module estimates the potential outcomes of unit $i$. The same colors denote the same latent roles \(u^i\), \(u^c\), and \(u^n\) across modules, while darker colors indicate the corresponding neighbor-aggregated features.
    }
    \label{fig: network}
\end{figure*}

\section{Methodology} \label{sec: method}

In this section, we devise our networked effect estimator in the presence of latent confounders.
Specifically, our estimator contains three modules, including the representation learning module, the feature module, and the outcome estimator module.
The representation learning module is built on Theorem \ref{theorem:recover latent}, aiming to correctly recover three types of latent confounders $u^i, u^c, u^n$.
The feature module is built on Theorems \ref{theorem: disentangle} and \ref{theorem: identify}, aggregating the information from units' and neighbors' information.
This aggregated information is then input into the outcome estimator module to predict the networked causal effects.
Overall, our model architecture is shown in Figure \ref{fig: network}.

\subsection{Representation Learning Module} \label{sec: Latent Variable Learning}

Following existing work \cite{guo2020learning,ma2021causal,jiang2022estimating, cai2023generalization,chen2024doubly}, we adopt Graph Convolutional Networks (GCN \cite{defferrard2016convolutional,kipf2016semi}) to encode the information of covariates of unit $i$ together with those of its neighbors, i.e., $x_i, x_ {\mathcal N_i}$:
\begin{equation}
\begin{aligned}
   & h^{neigh}_{i,1} =\sigma(\sum_{j\in \mathcal{N}_i}\frac{1}{\sqrt{d_id_j}}W_1^Tx_j), 
   \\& h_{i,2} = MLP_1(h^{neigh}_{i,1},x_i) ,
\end{aligned}
\end{equation}
where $\sigma(\cdot)$ denotes a nonlinear activation function, $d_i$ is the degree of unit $i$, $W_1$ is the learning weight matrix of the GCN, and $MLP_1$ is a multilayer perceptron (MLP). Here, the resulting representation $h_{i,2}$ serves as a representation of $x_i$ and $x_ {\mathcal N_i}$.

Then, given $x_i$ and $x_ {\mathcal N_i}$, we employ the identifiable representation learning technique to recover latent confounders.
Specifically, according to Assumption \ref{asmp: exp dist}, we parametrize the prior following \cite{lu2022invariant, zhu2025causal}:
\begin{equation}
  \begin{aligned}
   &p(u_i|x_{\mathcal{N}_i}) 
   \\= &
    \langle MLP_2(u_i), MLP_3(h_{i,2}) \rangle + \langle [u_i, u_i^2], MLP_4(h_{i,2}) \rangle
\end{aligned}
\end{equation}
where $u_i=[u_i^i, u_i^c, u_i^n]$ and $MLP_2(u_i)$ serves as $\bm{T}_{MLP}$,
the concatenated $[u_i, u_i^2]$ serves as $\bm{T}_f$,
$MLP_3(h_{i,2})$ serves as $\bm{\lambda}_{MLP}$,
and $ MLP_4(h_{i,2})$ serves as  $\bm{\lambda}_{f}$ in Assumption \ref{asmp: exp dist}.

As for the encoder, the variational approximation of the posterior is defined as:
\begin{equation}
  \begin{aligned}
    q(u^i_i,u^c_i,u^n_i|x_i,x_ {\mathcal N_i}) = \prod_{i=1}^{d_U} \mathcal{N}(\mu=\hat{\mu}_{u_{i}},\sigma^2=\hat{\sigma}_{u_{i}}^2),
\end{aligned}
\end{equation}
where $d_U$ is the dimension of $U$, and the parameters of the Gaussian distribution, the mean $\hat{\mu}_{u_{i}}=MLP_5(h_{i,2})$ and the variance $\hat{\sigma}_{u_{i}}^2=MLP_6(h_{i,2})$, are produced by two MLPs that take $h_{i,2}$ as input, with $MLP_6$ using a sigmoid activation to ensure positivity.

As for the decoder, for continuous proxies, we parametrize the probability distribution as a Gaussian distribution with its mean given by an MLP and a fixed variance $v^2$. For discrete proxies, we use a Bernoulli distribution parametrized by an MLP similarly:
\begin{equation}
  \begin{aligned}
    & p(x_i|u^i_i,u^c_i,u^n_i) = \prod_{i=1}^{d_X} \mathcal{N}(\mu=\hat{\mu}_x,\sigma^2=v_x^2) 
     \\
    \text{or}\hspace{4mm} & 
    p(x_i|u^i_i,u^c_i,u^n_i) =  \prod_{i=1}^{d_X} \mathbf{Bern}(\pi = \hat{\pi}_x), 
\end{aligned}
\end{equation}
where $d_X$ is the dimension of $X$, and for the continuous case, $\hat{\mu}_x$ is the mean of the Gaussian distribution parametrized by an MLP using the sampled $\hat u^i_i,\hat u^c_i,\hat u^n_i$ from the posterior as input, i.e., $\hat{\mu}_x=MLP_7(\hat u^i_i, \hat u^c_i, \hat u^n_i)$, and $v_x^2$ is the fixed variance of Gaussian distribution, and for the discrete case $\hat{\pi}_x$ is the mean of Bernoulli distribution similarly parametrized by an MLP taking $\hat u^i_i,\hat u^c_i,\hat u^n_i$ as input.

In this module, we use the negative variational Evidence Lower BOund (ELBO) as the loss function, defined as
\begin{equation} \label{ELBO}
  \begin{aligned}
    & \textbf{ELBO} =  \bE_{q(u^i,u^c,u^n|x , x_{\mathcal N})} [\log p(x|u^i,u^c,u^n) \\&
    + \log p(u^i,u^c,u^n|x_{\mathcal N})
    - \log q(u^i,u^c,u^n|x , x_{\mathcal N})].
\end{aligned}
\end{equation}
Detailed ELBO derivations are given in Appendix D. 

Since the prior belongs to a multivariate exponential family distribution with unknown normalization constant $\mathcal C$, it is infeasible to learn $\bm T$ and $\bm \lambda$ by optimizing the KL-divergence term in Eq. \eqref{ELBO}.
Therefore, we utilize the widely used score matching technique \cite{vincent2011connection} for training unnormalized densities to learn the parameters of $\bm T$ and $\bm \lambda$ by minimizing
\begin{equation}
  \begin{aligned}
    & \mathcal L_{sm} =    
    \bE_{q(u^i,u^c,u^n|x,x_{\mathcal N})} [
    \\& 
    \| \nabla _{u} \log q(u^i,u^c,u^n|x,x_{\mathcal N}) -\nabla_{u} \log p(u^i,u^c,u^n|x_{\mathcal N}) \|^2 ],
\end{aligned} 
\end{equation}
where $\nabla$ is the gradient operator. Therefore, the loss function for the representation learning module is 
\begin{equation}
  \begin{aligned}
    & \mathcal L_{rlm} =    
    \bE_{q(u^i,u^c,u^n|x,x_{\mathcal N})} [ - \log p(x|u^i,u^c,u^n) +
    \\& 
    \| \nabla _{u} \log q(u^i,u^c,u^n|x,x_{\mathcal N}) -\nabla_{u} \log p(u^i,u^c,u^n|x_{\mathcal N}) \|^2 ].
\end{aligned}
\end{equation}

\subsection{Feature Module and Outcome Estimator Module} \label{sec: Network Effect Estimator}

After obtaining sampled $\hat u^i, \hat u^c, \hat u^n$, we can aggregate the necessary information for the effect estimation.
The way of aggregation follows our Assumption \ref{asmp: Latent Network unconfounderness} and Figure \ref{fig: causal graph}. 
This means that, to predict the outcome, the necessary information is the confounders of the unit itself, including $u^i, u^c$, and the confounders of neighbors, including  $ u^c_{\mathcal N}, u^n_{\mathcal N}$.

Specifically, in the feature module, we first aggregate the neighbors' $\hat u^c, \hat u^n$ to obtain  $\hat u^c_{\mathcal N}, \hat u^n_{\mathcal N}$ via a GCN:
\begin{equation*}
\begin{aligned}
   & h_{i,3}^{neigh} =\sigma(\sum_{j\in \mathcal{N}_i}\frac{1}{\sqrt{d_id_j}}W_2^T [\hat u^c_j,\hat u^n_j]),    
   \\& h_{i,4} = MLP_{8}(h^{neigh}_{i,3}, \hat u^i_i, \hat u^c_i) ,
\end{aligned}
\end{equation*}
where $\sigma(\cdot)$ is a nonlinear activation function, $d_i$ is the degree of unit $i$, $W_2$ is the learning weight matrix of the GCN.
Here, $h_{i,3}^{neigh}$ serves as the neighbor latent confounders $[u^c_{\mathcal N_i}, u^n_{\mathcal N_i}]$ and $h_{i,4}$ serves as the whole confounders $[u^i_i, u^c_i, u^c_{\mathcal N_i}, u^n_{\mathcal N_i}]$, used to predict the outcomes $y_i$.

Then, we use $h_{i,4}$ and $t_i,z_i$ together to estimate $y_i$ of the treated and control groups respectively, i.e.,
\begin{equation} 
\mu^{NN}(t_i,z_i, x_i, x_{\mathcal N_i}) = 
\left\{
\begin{aligned}
MLP_{9}(z_i, h_{i,4}), & \quad  t_i =1  \\
MLP_{10}(z_i, h_{i,4}), & \quad  t_i =0 
\end{aligned}
\right.
,
\end{equation}
and the loss function is 
\begin{equation}
    \mathcal{L}_y = \Sigma_{i=1}^n  (y_i - \mu^{NN}(t_i,z_i, x_i, x_{\mathcal N_i}))^2.
\end{equation}
Moreover, inspired by \cite{jiang2022estimating,cai2023generalization}, we further consider a balancing regularization term to learn a balanced representation as our loss:
\begin{equation}
    \mathcal{L}_{IPM} =  \text{IPM} (p(h_{i,4},t_i,z_i), p(h_{i,4})p(t_i,z_i)),
\end{equation}
where $\text{IPM}(p, q)=\sup_{g\in\mathcal G} |\int_\mathcal{X} g(x)(p(x)-q(x))dx |$ is the integral probability metric, measuring the distance between two distributions $p,q$, which can be implemented by Wasserstein Distance.
Here the samples from $p(h_{i,4})p(t_i, z_i)$ are obtained by randomly permuting the observed treatment pairs $t_i$ and $z_i$.

Overall, our final loss function is 
\begin{equation}
    \mathcal{L}_{all}  =  \mathcal{L}_{rlm} +  \alpha \mathcal{L}_{IPM} + \beta \mathcal{L}_y ,
\end{equation}
where $\alpha$ and $\beta$ are the hyperparameters that control the weights of the respective loss terms.

We denote our method as \textbf{CaLaNet} (\textbf{Ca}usal representation learning-based \textbf{La}tent-confounder recovery for \textbf{Net}worked effect estimation).

\section{Experiments}

In this section, we assess the effectiveness of our proposed method, \textbf{CaLaNet}, and examine the validity of the underlying theoretical results using commonly adopted semisynthetic datasets. Specifically, we aim to address the following research questions (RQs):
\begin{itemize}
\item \textbf{RQ1:} How does the proposed method compare with existing approaches in terms of performance?
\item \textbf{RQ2:} Can the proposed model accurately recover the latent confounders?
\item \textbf{RQ3:} Does our model maintain stable performance under different hyperparameter settings?
\end{itemize}

\subsection{Experimental Setup}

To begin with, we first introduce our experimental setup, including the used datasets, the compared baselines, and the used metrics.

\subsubsection{Datasets}
 
 We consider two widely used semisynthetic datasets, BlogCatalog and Flickr, to verify the effectiveness of our estimator. We further use a synthetic dataset to validate the correctness of our theories, i.e., whether our method can correctly recover latent confounders.

Consistent with prior works \cite{jiang2022estimating, guo2020learning, ma2021deconfounding, chen2024doubly}, we evaluate our proposed method on two widely used semisynthetic datasets:
\begin{itemize} 
    \item \textbf{BlogCatalog (BC)}: an online blogging community where each node represents a blogger and each edge corresponds to a social connection between bloggers. Node features are constructed as bag-of-words representations derived from the keywords in blogger descriptions.
    \item \textbf{Flickr}: a social media platform for sharing images and videos, where nodes correspond to users and edges denote social relationships. Node features are based on users’ tag lists reflecting their interests.
\end{itemize}

We reuse the original covariates as the latent confounders and then divide them into $u^i, u^c, u^n$.
We generate the proxies $x_i$ using 
$x_i = w_1 u_i + e_i$ where $w_1 $ are randomly sampled from Uniform distribution $\mathcal{U}(0.5, 1)$ and $e_{i}$ is standard Gaussian noise. Then given the latent confounder $u^i_i, u^c_i, u^n_i$ of unit $i$, the treatment mechanism, consistent with \cite{jiang2022estimating,cai2023generalization,chen2024doubly}, is defined as
\begin{equation*}
    \begin{aligned}
        t_i=
            \begin{cases}
            1& \text{if \quad $tpt_i>\overline{tpt}$},\\
            0& \text{else},
            \end{cases}
    \end{aligned}
\end{equation*}
where $\overline{tpt}$ is obtained by averaging all $tpt_i$, and $tpt_i = pt_i + pt_{\mathcal N_i}$.
Here, $pt_i = Sigmoid(w_2 \times [u^i_i, u^c_i])$, and $pt_{\mathcal N _i}= \frac{1}{|\mathcal N_i|} \sum_j^{j\in \mathcal N_i} Sigmoid(w_3 \times [u^c_{j}, u^n_{j}]) $ with $pt_i$ representing the unit-level treatment mechanism and $pt_{\mathcal N_i}$ capturing the influence of neighbors.
The weight vectors $w_2$ and $w_3$ are randomly generated to mimic the causal mechanism from the latent confounders to treatments. 
Finally, the neighborhood exposure $z_i$ can be directly obtained by the network topology $E$ and $t_{\mathcal N_i}$.

We then modify the data generation of outcome $y$ in \cite{jiang2022estimating} by replacing the original covariates $x_i$ with the latent confounders $u_i$. Specifically, the outcome is generated as:
\begin{equation*}
    \begin{aligned}
        y_i(t_i,z_i) = t_i + z_i + po_i + 0.5 \times po_{\mathcal N_i} + e_{y,i},
    \end{aligned}
\end{equation*}
where $e_{y,i}$ denotes Gaussian noise.
Here, $po_i = Sigmoid(w_4 \times u_i+w_5 \times u_c)$, and $ po_{\mathcal N_i}$ is the average of $Sigmoid(w_6 \times u_c +w_7 \times u_n)$. 
$w_4,w_5,w_6$, and $w_7$ are all randomly generated weight vectors that mimic the causal mechanism from the confounders to outcomes.
We denote these datasets as \textbf{BC(homo)} and \textbf{Flickr(homo)}\footnote{Original datasets are available at \url{https://github.com/songjiang0909/Causal-Inference-on-Networked-Data}.} since this generation of $y$ only measures the homogeneous causal effects.

Also following \cite{chen2024doubly}, we consider the data generation of outcome $y$ with heterogeneous effects:
\begin{equation*}
    \begin{aligned}
         y_i& (t_i,z_i) = t_i + z_i + po_i + 0.5 \times po_{\mathcal N_i} 
        \\& + t_i (po_i+0.5\times po_{\mathcal N_i})
        +  t_i (0.5 \times po_i+ po_{\mathcal N_i})
        + e_{y,i},
    \end{aligned}
\end{equation*}
and the datasets are denoted as \textbf{BC(hete)} and \textbf{Flickr(hete)}.

Due to space limitation, we leave the detailed data generation process of the synthetic dataset in Appendix E.

\begin{table*}[!h] 
\renewcommand{\arraystretch}{1.7}
\caption{Experimental results on BC(homo) Dataset. The top result is highlighted in bold, and the runner-up is underlined.}     
\label{tab: BC}
\centering 
\resizebox{\linewidth}{!}{ \huge
\begin{tabular}{@{}l|ccc| ccc | ccc| ccc @{}} 
\hline
 & \multicolumn{6}{c|}{$\varepsilon_{average}$}  & \multicolumn{6}{c}{$\varepsilon_{individual}$}  \\ \hline
 & \multicolumn{3}{c|}{Within Sample}
 & \multicolumn{3}{c|}{Out-of Sample}
 & \multicolumn{3}{c|}{Within Sample}
 & \multicolumn{3}{c}{Out-of Sample}  \\ \hline
 Methods & AME & ASE  & ATE & AME & ASE  & ATE &    
 IME & ISE  & ITE &  IME & ISE  & ITE \\ \hline
TARNET+z  &  { $0.1573 _{\pm 0.0405} $} &  { $0.0824 _{\pm0.0149} $} &  { $0.2046 _{\pm0.0193} $} &  { $0.1492 _{\pm 0.0370} $} &  { $0.0855 _{\pm0.0147} $} &  { $0.1982 _{\pm0.0380} $} &  { $0.2096 _{\pm 0.0250} $} &  { $0.1161 _{\pm0.0159} $} &  { $0.2444 _{\pm0.0239} $} &  { $0.2809 _{\pm 0.0507} $} &  { $0.1209 _{\pm0.0152} $} &  { $0.3062 _{\pm0.0627} $}   \\ \hline
CFR+z  &  { $\underline{0.0788} _{\pm 0.0096} $} &  { $0.1157 _{\pm0.0076} $} &  { $0.2323 _{\pm0.0106} $} &  { $\underline{0.0770} _{\pm 0.0099} $} &  { $0.1157 _{\pm0.0075} $} &  { $0.2306 _{\pm0.0106} $} &  { $\underline{0.0796} _{\pm 0.0091} $} &  { $0.1158 _{\pm0.0076} $} &  { $0.2325 _{\pm0.0106} $} &  { $0.1000 _{\pm 0.0388} $} &  { $0.1157 _{\pm0.0075} $} &  { $0.2405 _{\pm0.0166} $}  \\ \hline
GEst &  { $0.1872 _{\pm 0.0672} $} &  { $0.2369 _{\pm0.0607} $} &  { $0.1422 _{\pm0.0562} $} &  { $0.1955 _{\pm 0.0611} $} &  { $0.2391 _{\pm0.0617} $} &  { $0.1302 _{\pm0.0524} $} &  { $0.2307 _{\pm 0.0493} $} &  { $0.2603 _{\pm0.0586} $} &  { $0.1877 _{\pm0.0495} $} &  { $0.2388 _{\pm 0.0431} $} &  { $0.2623 _{\pm0.0592} $} &  { $0.1790 _{\pm0.0440} $}   \\ \hline
ND+z   &  { $0.2375 _{\pm 0.0450} $} &  { $\underline{0.0316} _{\pm0.0104} $} &  { $0.0790 _{\pm0.0226} $} &  { $0.2380 _{\pm 0.0458} $} &  { $\underline{0.0323} _{\pm0.0122} $} &  { $0.0768 _{\pm0.0254} $} &  { $0.2377 _{\pm 0.0448} $} &  { $\underline{0.0321} _{\pm0.0101} $} &  { $0.0792 _{\pm0.0226} $} &  { $0.2477 _{\pm 0.0460} $} &  { $\underline{0.0379} _{\pm0.0099} $} &  { $0.1068 _{\pm0.0172} $} 
 \\ \hline
NetEst  &  { $0.0989 _{\pm 0.0735} $} &  { ${0.0428} _{\pm0.0487} $} &  { $0.0596 _{\pm0.0291} $} &  { $0.0995 _{\pm 0.0725} $} &  { $0.0425 _{\pm0.0487} $} &  { $0.0599 _{\pm0.0286} $} &  { $0.0992 _{\pm 0.0732} $} &  { $0.0431 _{\pm0.0485} $} &  { $0.0602 _{\pm0.0290} $} &  { $0.1130 _{\pm 0.0617} $} &  { $0.0553 _{\pm0.0443} $} &  { $0.0660 _{\pm0.0254} $}   \\ \hline
RRNet &   { $0.0884 _{\pm 0.0495} $} &  { $0.0445 _{\pm0.0158} $} &  { $0.0768 _{\pm0.0239} $} &  { $0.0892 _{\pm 0.0505} $} &  { $0.0447 _{\pm0.0160} $} &  { $0.0782 _{\pm0.0253} $} &  { $0.0915 _{\pm 0.0462} $} &  { $0.0452 _{\pm0.0153} $} &  { $0.0865 _{\pm0.0260} $} &  { $\underline{0.0917} _{\pm 0.0491} $} &  { $0.0453 _{\pm0.0159} $} &  { $0.0887 _{\pm0.0282} $}  \\ \hline
SPNet+z &   { $ 0.0802 _{\pm 0.0632} $} &  { $0.0446 _{\pm0.0155} $} &  { $0.0570 _{\pm0.0188} $} &  { $0.0931 _{\pm 0.0607} $} &  { $0.0394 _{\pm0.0215} $} &  { $\underline{0.0385} _{\pm0.0183} $} &  { $0.1153 _{\pm 0.0482} $} &  { $0.0741 _{\pm0.0109} $} &  { $0.1094 _{\pm0.0274} $} &  { $0.1358 _{\pm 0.0566} $} &  { $0.0604 _{\pm0.0145} $} &  { $0.0933 _{\pm0.0342} $}  \\ \hline
TNet  & { $0.1045 _{\pm 0.0610} $} &  { $0.0502 _{\pm0.0559} $} &  { $\underline{0.0473} _{\pm0.0229} $} &  { $0.1045 _{\pm 0.0610} $} &  { $0.0502 _{\pm0.0559} $} &  { $0.0473 _{\pm0.0229} $} &  { $0.1045 _{\pm 0.0610} $} &  { $0.0502 _{\pm0.0559} $} &  { $\underline{0.0473} _{\pm0.0229} $} &  { $0.1045 _{\pm 0.0610} $} &  { $0.0502 _{\pm0.0559} $} &  { $\underline{0.0473} _{\pm0.0229} $}  \\ \hline
CaLaNet &  { $\textbf{0.0551} _{\pm 0.0423} $} &  { $\textbf {0.0279} _{\pm0.0279} $} &  { $\textbf{0.0288} _{\pm0.0150} $} &  { $\textbf{0.0549} _{\pm 0.0426} $} &  { $\textbf{0.0278} _{\pm0.0279} $} &  { $\textbf{0.0288} _{\pm0.0150} $} &  { $\textbf{0.0554} _{\pm 0.0418} $} &  { $\textbf{0.0279} _{\pm0.0279} $} &  { $\textbf{0.0289} _{\pm0.0149} $} &  { $\textbf{0.0551} _{\pm 0.0424} $} &  { $\textbf{0.0278} _{\pm0.0279} $} &  { $\textbf{0.0288} _{\pm0.0149} $}  \\ \hline
\end{tabular}
}
\end{table*}

\begin{table*}[!h] 
\renewcommand{\arraystretch}{1.7}
\caption{Experimental results on BC(hete) Dataset. The top result is highlighted in bold, and the runner-up is underlined.}     
\label{tab: BC_hete}
\centering 
\resizebox{\linewidth}{!}{ \huge
\begin{tabular}{@{}l|ccc| ccc | ccc| ccc @{}} 
        \hline
         & \multicolumn{6}{c|}{$\varepsilon_{average}$}  & \multicolumn{6}{c}{$\varepsilon_{individual}$}  \\ \hline
         & \multicolumn{3}{c|}{Within Sample}
         & \multicolumn{3}{c|}{Out-of Sample}
         & \multicolumn{3}{c|}{Within Sample}
         & \multicolumn{3}{c}{Out-of Sample}  \\ \hline
         Methods & AME & ASE  & ATE & AME & ASE  & ATE &    
         IME & ISE  & ITE &  IME & ISE  & ITE \\ \hline
        TARNET+z  & { $0.2538 _{\pm 0.1127} $} &  { $0.1657 _{\pm0.0563} $} &  { $0.3866 _{\pm0.0711} $} &  { $0.2619 _{\pm 0.1054} $} &  { $0.1701 _{\pm0.0594} $} &  { $0.4044 _{\pm0.1334} $} &  { $0.3455 _{\pm 0.0654} $} &  { $0.2122 _{\pm0.0558} $} &  { $0.4605 _{\pm0.0720} $} &  { $0.5590 _{\pm 0.2643} $} &  { $0.2207 _{\pm0.0510} $} &  { $0.6349 _{\pm0.3280} $}  \\ \hline
        CFR+z  & { $0.1580 _{\pm 0.0189} $} &  { $0.2071 _{\pm0.0237} $} &  { $0.4067 _{\pm0.0407} $} &  { $0.1559 _{\pm 0.0203} $} &  { $0.2076 _{\pm0.0245} $} &  { $0.4061 _{\pm0.0449} $} &  { $0.1825 _{\pm 0.0129} $} &  { $0.2092 _{\pm0.0233} $} &  { $0.4316 _{\pm0.0351} $} &  { $0.2058 _{\pm 0.0432} $} &  { $0.2098 _{\pm0.0242} $} &  { $0.4422 _{\pm0.0456} $} \\ \hline
        GEst   & { $0.2734 _{\pm 0.1240} $} &  { $0.4257 _{\pm0.0973} $} &  { $0.2916 _{\pm0.1119} $} &  { $0.2722 _{\pm 0.1308} $} &  { $0.4277 _{\pm0.1012} $} &  { $0.2873 _{\pm0.1220} $} &  { $0.3334 _{\pm 0.1082} $} &  { $0.4592 _{\pm0.0934} $} &  { $0.3546 _{\pm0.0947} $} &  { $0.3832 _{\pm 0.1474} $} &  { $0.4612 _{\pm0.0972} $} &  { $0.3958 _{\pm0.1486} $}  \\ \hline
        ND+z  &  { $0.4124 _{\pm 0.0702} $} &  { $\underline{0.0451} _{\pm0.0201} $} &  { $0.1330 _{\pm0.0205} $} &  { $0.4111 _{\pm 0.0737} $} &  { $0.0486 _{\pm0.0206} $} &  { $0.1326 _{\pm0.0261} $} &  { $0.4226 _{\pm 0.0673} $} &  { $\underline{0.0562} _{\pm0.0146} $} &  { $0.1941 _{\pm0.0246} $} &  { $0.4330 _{\pm 0.0662} $} &  { $0.0666 _{\pm0.0137} $} &  { $0.2211 _{\pm0.0321} $} \\ \hline
        NetEst  & { $0.2301 _{\pm 0.0784} $} &  { $0.0799 _{\pm0.0453} $} &  { $0.1296 _{\pm0.0594} $} &  { $0.2252 _{\pm 0.0729} $} &  { $0.0785 _{\pm0.0441} $} &  { $0.1221 _{\pm0.0594} $} &  { $0.2479 _{\pm 0.0737} $} &  { $0.0859 _{\pm0.0433} $} &  { $0.1960 _{\pm0.0444} $} &  { $0.2485 _{\pm 0.0761} $} &  { $0.0923 _{\pm0.0433} $} &  { $0.1988 _{\pm0.0435} $}  \\ \hline
        RRNet &   { $0.1628 _{\pm 0.1183} $} &  { ${0.0476 }_{\pm0.0416} $} &  { $0.0919 _{\pm0.0581} $} &  { $0.1636 _{\pm 0.1162} $} &  { $ \underline{0.0474 }_{\pm0.0424} $} &  { $0.0961 _{\pm0.0640} $} &  { $0.1967 _{\pm 0.0979} $} &  { $0.0591 _{\pm0.0364} $} &  { $0.1719 _{\pm0.0485} $} &  { $0.1974 _{\pm 0.0960} $} &  { $\underline{0.0595} _{\pm0.0365} $} &  { $0.1764 _{\pm0.0505} $} \\ \hline
        SPNet+z &  { $\underline{0.0706} _{\pm 0.0738} $} &  { $0.0714 _{\pm0.0256} $} &  { $0.1169 _{\pm0.0317} $} &  { $\underline{0.0805} _{\pm 0.0716} $} &  { $0.0671 _{\pm0.0230} $} &   $0.0841 _{\pm0.0419} $ &  { $0.1743 _{\pm 0.0338} $} &  { $0.1054 _{\pm0.0138} $} &  { $0.2209 _{\pm0.0288} $} &  { $\underline{0.1901} _{\pm 0.0480} $} &  { $0.0906 _{\pm0.0190} $} &  { $0.2012 _{\pm0.0480} $} \\ \hline
        TNet  & { $0.1216 _{\pm 0.0864} $} &  { $0.0537 _{\pm0.0524} $} &  { $\underline{0.0429} _{\pm0.0301} $} &  { $0.1257 _{\pm 0.0727} $} &  { $0.0537 _{\pm0.0511} $} &  { $0.0481 _{\pm0.0269} $} &  { $\underline{0.1731} _{\pm 0.0450} $} &  { $0.0655 _{\pm0.0465} $} &  { $\textbf{0.1458} _{\pm0.0175} $} &  { $0.1915 _{\pm 0.0542} $} &  { $0.0650 _{\pm0.0458} $} &  { $\underline{0.1740} _{\pm0.0621} $}  \\ \hline
        CaLaNet &  { $\textbf{0.0701} _{\pm 0.0571} $} &  { $\textbf{0.0321} _{\pm0.0239} $} &  { $\textbf{0.0401} _{\pm0.0264} $} &  { $\textbf{0.0691} _{\pm 0.0550} $} &  { $\textbf{0.0305} _{\pm0.0243} $} &  { $\textbf{0.0396} _{\pm0.0265} $} &  { $\textbf{0.1224} _{\pm 0.0310} $} &  { $\textbf{0.0458} _{\pm0.0181} $} &  { $\underline{0.1481} _{\pm0.0210} $} &  { $\textbf{0.1216} _{\pm 0.0300} $} &  { $\textbf{0.0452} _{\pm0.0178} $} &  { $\textbf{0.1493} _{\pm0.0185} $} 
        \\ \hline
    \end{tabular}
    }
\end{table*}

\begin{table*}[!t] 
\renewcommand{\arraystretch}{1.7}
\caption{Experimental results on Flickr(homo) Dataset. The top result is highlighted in bold, and the runner-up is underlined.}     
\label{tab: Flickr}
\centering 
\resizebox{\linewidth}{!}{ \huge
\begin{tabular}{@{}l|ccc| ccc | ccc| ccc @{}} 
        \hline
         & \multicolumn{6}{c|}{$\varepsilon_{average}$}  & \multicolumn{6}{c}{$\varepsilon_{individual}$}  \\ \hline
         & \multicolumn{3}{c|}{Within Sample}
         & \multicolumn{3}{c|}{Out-of Sample}
         & \multicolumn{3}{c|}{Within Sample}
         & \multicolumn{3}{c}{Out-of Sample}  \\ \hline
         Methods & AME & ASE  & ATE & AME & ASE  & ATE &    
         IME & ISE  & ITE &  IME & ISE  & ITE \\ \hline
        TARNET+z  & { $0.0783 _{\pm 0.0418} $} &  { $0.0874 _{\pm0.0213} $} &  { $0.2025 _{\pm0.0396} $} &  { $0.0976 _{\pm 0.0506} $} &  { $0.0724 _{\pm0.0184} $} &  { $0.1356 _{\pm0.0587} $} &  { $0.1362 _{\pm 0.0254} $} &  { $0.1103 _{\pm0.0194} $} &  { $0.2358 _{\pm0.0383} $} &  { $1.0869 _{\pm 1.2258} $} &  { $0.1011 _{\pm0.0185} $} &  { $1.0889 _{\pm1.2270} $}  \\ \hline
        CFR+z  &   { $0.0579 _{\pm 0.0247} $} &  { $0.0785 _{\pm0.0070} $} &  { $0.1651 _{\pm0.0121} $} &  { $0.0507 _{\pm 0.0192} $} &  { $0.0783 _{\pm0.0070} $} &  { $0.1581 _{\pm0.0097} $} &  { $0.0599 _{\pm 0.0240} $} &  { $0.0786 _{\pm0.0070} $} &  { $0.1654 _{\pm0.0120} $} &  { $0.3465 _{\pm 0.4615} $} &  { $0.0786 _{\pm0.0069} $} &  { $0.4102 _{\pm0.4278} $} \\ \hline
        GEst   & { $0.1551 _{\pm 0.0130} $} &  { $0.2475 _{\pm0.0476} $} &  { $0.0805 _{\pm0.0325} $} &  { $0.1511 _{\pm 0.0137} $} &  { $0.2494 _{\pm0.0470} $} &  { $0.0805 _{\pm0.0278} $} &  { $0.1779 _{\pm 0.0122} $} &  { $0.2656 _{\pm0.0378} $} &  { $0.1268 _{\pm0.0160} $} &  { $0.2867 _{\pm 0.2172} $} &  { $0.2677 _{\pm0.0372} $} &  { $0.2471 _{\pm0.2352} $}  \\ \hline
        ND+z  &  { $0.1416 _{\pm 0.0240} $} &  { $\underline{0.0204} _{\pm0.0093} $} &  { $0.0478 _{\pm0.0216} $} &  { $0.1435 _{\pm 0.0364} $} &  { $\underline{0.0226} _{\pm0.0101} $} &  { $0.0485 _{\pm0.0236} $} &  { $0.1427 _{\pm 0.0246} $} &  { $\underline{0.0221} _{\pm0.0090} $} &  { $0.0501 _{\pm0.0178} $} &  { $0.3849 _{\pm 0.2395} $} &  { $0.0348 _{\pm0.0078} $} &  { $0.3453 _{\pm0.2772} $}  \\ \hline
        NetEst  & { $0.0515 _{\pm 0.0538} $} &  { $0.0355 _{\pm0.0317} $} &  { $0.0715 _{\pm0.0381} $} &  { $0.0470 _{\pm 0.0500} $} &  { $0.0338 _{\pm0.0330} $} &  { $0.0529 _{\pm0.0395} $} &  { $0.0844 _{\pm 0.0406} $} &  { $0.0566 _{\pm0.0253} $} &  { $0.1043 _{\pm0.0312} $} &  { $0.2934 _{\pm 0.3001} $} &  { $0.2809 _{\pm0.3387} $} &  { $0.3068 _{\pm0.1860} $} \\ \hline
        RRNet &  { $ \underline{ 0.0296} _{\pm 0.0123} $} &  { ${0.0251} _{\pm0.0172} $} &  { $ \underline{0.0199} _{\pm0.0179} $} &  { $ \underline{0.0296} _{\pm 0.0123} $} &  { $ {0.0251} _{\pm0.0172} $} &  { $\underline{0.0199} _{\pm0.0179} $} &  { $ \underline{0.0296} _{\pm 0.0123} $} &  { ${0.0251} _{\pm0.0172} $} &  { $\underline{0.0199} _{\pm0.0179} $} &  { $ \underline{0.0296} _{\pm 0.0123} $} &  { $ \underline{0.0251} _{\pm0.0172} $} &  { $\underline{0.0199} _{\pm0.0179} $}  \\ \hline
        SPNet+z &  { $0.0366 _{\pm 0.0359} $} &  { $0.0604 _{\pm0.0371} $} &  { $0.1267 _{\pm0.0655} $} &  { $0.0451 _{\pm 0.0332} $} &  { $0.0445 _{\pm0.0300} $} &  { $0.0865 _{\pm0.0410} $} &  { $0.0639 _{\pm 0.0273} $} &  { $0.0980 _{\pm0.0269} $} &  { $0.1877 _{\pm0.0546} $} &  { $0.0834 _{\pm 0.0342} $} &  { $0.0694 _{\pm0.0234} $} &  { $0.1309 _{\pm0.0500} $}  \\ \hline
        TNet  & { $0.0319 _{\pm 0.0249} $} &  { $0.0274 _{\pm0.0309} $} &  { $0.0735 _{\pm0.0240} $} &  { $0.0299 _{\pm 0.0231} $} &  { $0.0277 _{\pm0.0313} $} &  { $0.0715 _{\pm0.0214} $} &  { $0.0347 _{\pm 0.0282} $} &  { $0.0276 _{\pm0.0313} $} &  { $0.0752 _{\pm0.0263} $} &  { $0.0561 _{\pm 0.0648} $} &  { $0.0286 _{\pm0.0331} $} &  { $0.0918 _{\pm0.0555} $}  \\ \hline
        CaLaNet &   { $\textbf{0.0223} _{\pm 0.0159} $} &  { $\textbf{0.0147} _{\pm0.0068} $} &  { $\textbf{0.0151} _{\pm0.0127} $} &  { $\textbf{0.0222} _{\pm 0.0158} $} &  { $ \textbf{0.0148} _{\pm0.0069} $} &  { $\textbf{0.0154} _{\pm0.0128} $} &  { $\textbf{0.0223} _{\pm 0.0159} $} &  { $\textbf{0.0149} _{\pm0.0069} $} &  { $ \textbf{0.0152} _{\pm0.0127} $} &  { $\textbf{0.0223} _{\pm 0.0158} $} &  { $\textbf{0.0148} _{\pm0.0068} $} &  { $\textbf{0.0155} _{\pm0.0129} $} 
        \\ \hline
    \end{tabular}
    }
\end{table*}

\subsubsection{Baselines} 

We denote our method as \textbf{CaLaNet} \footnote{Our code is available at \url{https://github.com/WeilinChen507/CaLaNet} and \url{https://github.com/DMIRLAB-Group/CaLaNet}.}. 
We compare CaLaNet against several state-of-the-art baselines, which can be grouped into (i) adaptations of no-interference methods and (ii) methods designed for networked interference:

\begin{itemize} 
    \item \textbf{TARNET+z}: The original TARNET \cite{johansson2021generalization} employs a two-headed neural network, similar to a T-learner, to estimate causal effects under the no-interference assumption. We extend TARNET by incorporating neighborhood exposure $z_i$ as an additional input.
    \item \textbf{CFR+z}: The original CFR \cite{johansson2021generalization} also adopts a two-headed neural network, augmented with a Maximum Mean Discrepancy (MMD) term to enforce balanced representations for counterfactual regression under no-interference. We adapt CFR by additionally inputting $z_i$.
    \item \textbf{ND+z}: The Network Deconfounder (ND) framework \cite{guo2020learning} leverages network information to address confounding under the no-interference assumption. We modify ND by adding $z_i$ as an input.
    \item \textbf{GEst} \cite{ma2021causal}: GEst builds on CFR by incorporating a Graph Convolutional Network (GCN) to aggregate neighbors’ features, while also including $z_i$, to estimate causal effects under networked interference.
    \item \textbf{NetEst} \cite{jiang2022estimating}: NetEst learns balanced representations via adversarial training for networked causal effect estimation.
    \item \textbf{RRNet} \cite{cai2023generalization}: RRNet combines the representation learning and reweighting techniques to estimate causal effects under interference. 
    \item \textbf{SPNet+z} \cite{huang2023modeling}: SPNet is designed to estimate the effect of the individual treatment $t_i$ while holding $z_i$ fixed. We adapt it to our setting by explicitly incorporating $z_i$ as an additional input.
    \item \textbf{TNet} \cite{chen2024doubly}: TNet utilizes targeted learning techniques in its neural network model to estimate causal effects under networked interference in a double robust manner. 
\end{itemize}

\begin{table*}[!t] 
\renewcommand{\arraystretch}{1.7}
\caption{Experimental results on Flickr(hete) Dataset. The top result is highlighted in bold, and the runner-up is underlined.}     
\label{tab: Flickr_hete}
\centering 
\resizebox{\linewidth}{!}{ \huge
\begin{tabular}{@{}l|ccc| ccc | ccc| ccc @{}} 
        \hline
         & \multicolumn{6}{c|}{$\varepsilon_{average}$}  & \multicolumn{6}{c}{$\varepsilon_{individual}$}  \\ \hline
         & \multicolumn{3}{c|}{Within Sample}
         & \multicolumn{3}{c|}{Out-of Sample}
         & \multicolumn{3}{c|}{Within Sample}
         & \multicolumn{3}{c}{Out-of Sample}  \\ \hline
         Methods & AME & ASE  & ATE & AME & ASE  & ATE &    
         IME & ISE  & ITE &  IME & ISE  & ITE \\ \hline
        TARNET+z  & { $0.1315 _{\pm 0.0740} $} &  { $0.1673 _{\pm0.0423} $} &  { $0.3590 _{\pm0.0785} $} &  { $0.1554 _{\pm 0.1110} $} &  { $0.1319 _{\pm0.0307} $} &  { $0.2728 _{\pm0.1321} $} &  { $0.2320 _{\pm 0.0432} $} &  { $0.2042 _{\pm0.0479} $} &  { $0.4254 _{\pm0.0802} $} &  { $1.5274 _{\pm 1.6256} $} &  { $0.1760 _{\pm0.0351} $} &  { $1.5957 _{\pm1.5779} $}  \\ \hline
        CFR+z  &  { $0.1131 _{\pm 0.0476} $} &  { $0.1437 _{\pm0.0081} $} &  { $0.2960 _{\pm0.0182} $} &  { $0.0998 _{\pm 0.0458} $} &  { $0.1412 _{\pm0.0085} $} &  { $0.2789 _{\pm0.0309} $} &  { $0.1445 _{\pm 0.0407} $} &  { $0.1463 _{\pm0.0083} $} &  { $0.3242 _{\pm0.0224} $} &  { $0.5946 _{\pm 0.7379} $} &  { $0.1448 _{\pm0.0087} $} &  { $0.7104 _{\pm0.6761} $} \\ \hline
        GEst   & { $0.3283 _{\pm 0.0426} $} &  { $0.4717 _{\pm0.1336} $} &  { $0.1074 _{\pm0.0255} $} &  { $0.3356 _{\pm 0.0270} $} &  { $0.4723 _{\pm0.1312} $} &  { $0.0969 _{\pm0.0099} $} &  { $0.3697 _{\pm 0.0386} $} &  { $0.5123 _{\pm0.1231} $} &  { $0.2178 _{\pm0.0214} $} &  { $0.7124 _{\pm 0.6463} $} &  { $0.5144 _{\pm0.1202} $} &  { $0.5914 _{\pm0.7073} $}  \\ \hline
        ND+z  &  { $0.2420 _{\pm 0.0330} $} &  { $0.0293 _{\pm0.0113} $} &  { $0.0852 _{\pm0.0365} $} &  { $0.2433 _{\pm 0.0539} $} &  { $0.0318 _{\pm0.0134} $} &  { $0.0785 _{\pm0.0422} $} &  { $0.2571 _{\pm 0.0348} $} &  { $0.0430 _{\pm0.0040} $} &  { $0.1607 _{\pm0.0188} $} &  { $0.5156 _{\pm 0.2180} $} &  { $0.0669 _{\pm0.0059} $} &  { $0.4720 _{\pm0.2580} $}  \\ \hline
        NetEst  & { $0.0530 _{\pm 0.0423} $} &  { $0.0452 _{\pm0.0351} $} &  { $0.0723 _{\pm0.0319} $} &  { $0.0466 _{\pm 0.0322} $} &  { $0.0565 _{\pm0.0454} $} &  { $0.0818 _{\pm0.0379} $} &  { $0.1145 _{\pm 0.0278} $} &  { $0.0667 _{\pm0.0267} $} &  { $0.1660 _{\pm0.0163} $} &  { $0.6855 _{\pm 0.2607} $} &  { $0.5353 _{\pm0.2507} $} &  { $0.5625 _{\pm0.1367} $}  \\ \hline
        RRNet &   { $0.0441 _{\pm 0.0196} $} &  { ${0.0280} _{\pm0.0153} $} &  { $\underline{0.0367} _{\pm0.0243} $} &  { $0.0446 _{\pm 0.0183} $} &  { $0.0292 _{\pm0.0165} $} &  { ${0.0329} _{\pm0.0255} $} &  { $0.0940 _{\pm 0.0166} $} &  { $0.0401 _{\pm0.0105} $} &  { $0.1373 _{\pm0.0188} $} &  { $0.0952 _{\pm 0.0153} $} &  { $0.0436 _{\pm0.0100} $} &  { $0.1419 _{\pm0.0202} $}    \\ \hline
        SPNet+z &  { $0.0429 _{\pm 0.0461} $} &  { $0.0885 _{\pm0.0509} $} &  { $0.2359 _{\pm0.0724} $} &  { $0.0509 _{\pm 0.0404} $} &  { $0.0555 _{\pm0.0264} $} &  { $0.1483 _{\pm0.0573} $} &  { $0.1296 _{\pm 0.0241} $} &  { $0.1593 _{\pm0.0314} $} &  { $0.3538 _{\pm0.0513} $} &  { $0.1827 _{\pm 0.1051} $} &  { $0.1053 _{\pm0.0208} $} &  { $0.2879 _{\pm0.0823} $}  \\ \hline
        TNet  & { $\underline{0.0411} _{\pm 0.0238} $} &  { $\underline{0.0206} _{\pm0.0073} $} &  { $\underline{0.0282} _{\pm0.0297} $} &  { $\underline{0.0417} _{\pm 0.0237} $} &  { $\underline{0.0196} _{\pm0.0098} $} &  { $\underline{0.0268} _{\pm0.0314} $} &  { $\underline{0.0936} _{\pm 0.0170} $} &  { $\underline{0.0338} _{\pm0.0065} $} &  { $\underline{0.1360} _{\pm0.0210} $} &  { $\underline{0.0950} _{\pm 0.0157} $} &  { $\underline{0.0361} _{\pm0.0074} $} &  { $\underline{0.1415} _{\pm0.0223} $}  \\ \hline
        CaLaNet &  { $\textbf{0.0283} _{\pm 0.0223} $} &  { $\textbf{0.0135} _{\pm0.0070} $} &  { $\textbf{0.0257} _{\pm0.0099} $} &  { $\textbf{0.0316} _{\pm 0.0206} $} &  { $\textbf{0.0146} _{\pm0.0074} $} &  { $\textbf{0.0217} _{\pm0.0091} $} &  { $\textbf{0.0892} _{\pm 0.0111} $} &  { $\textbf{0.0302} _{\pm0.0053} $} &  { $\textbf{0.1334} _{\pm0.0144} $} &  { $\textbf{0.0911} _{\pm 0.0099} $} &  { $\textbf{0.0334} _{\pm0.0055} $} &  { $\textbf{0.1383} _{\pm0.0151} $} 
        \\ \hline
    \end{tabular}
    }
\end{table*}

\subsubsection{Metrics}

In this paper, we evaluate performance using two metrics.
For average effects (AME, ASE, and ATE), we report the Mean Absolute Error (MAE): $\varepsilon_{average}= | \hat \tau - \tau|$, where $\tau$ and $\hat \tau$ are the average causal effect and the estimated one, respectively.
For individual-level effects (IME, ISE, and ITE), we adopt the Root Precision in Estimation of Heterogeneous Effect: $\varepsilon_{individual} = \sqrt{\frac{1}{n} \Sigma_{i=1}^n (\hat \tau_i - \tau_i)^2 }$, where $\tau_i$ and $\hat{\tau}_i$ represent the true and the estimated individual causal effects of unit $i$.
All results are reported as the mean and standard deviation over five independent runs.

\subsection{Experimental Analyses}

\subsubsection{\textbf{RQ1: Effectiveness of CaLaNet}}

As shown in Tables \ref{tab: BC}, \ref{tab: BC_hete}, \ref{tab: Flickr}, and \ref{tab: Flickr_hete}, we have conducted experiments by running our proposed method, CaLaNet, and several baselines. Overall, our CaLaNet outperforms all methods consistently with smaller estimation errors in all metrics, indicating the effectiveness of our method.
Specifically, compared with the baselines, our method performs better in terms of both average and heterogeneous treatment effect estimation, with smaller errors and standard deviations.
This means our CaLaNet is not only accurate for networked effect estimation but also performs stably.
This is reasonable since most existing methods do not consider the latent confounders that hinder the effect identification, while our method utilizes representation learning techniques and thereby achieves superior performance with recovered latent confounders.
In particular, SPNet+z also considers latent confounders. However, it has no theoretical guarantee for the latent confounder recovery, and thus underperforms our CaLaNet.
Hence, the results in Tables \ref{tab: BC}, \ref{tab: BC_hete},  \ref{tab: Flickr}, and \ref{tab: Flickr_hete} showcase the effectiveness of our proposed CaLaNet.

\begin{figure}[!h]
    \centering
    \includegraphics[width=1.\linewidth]{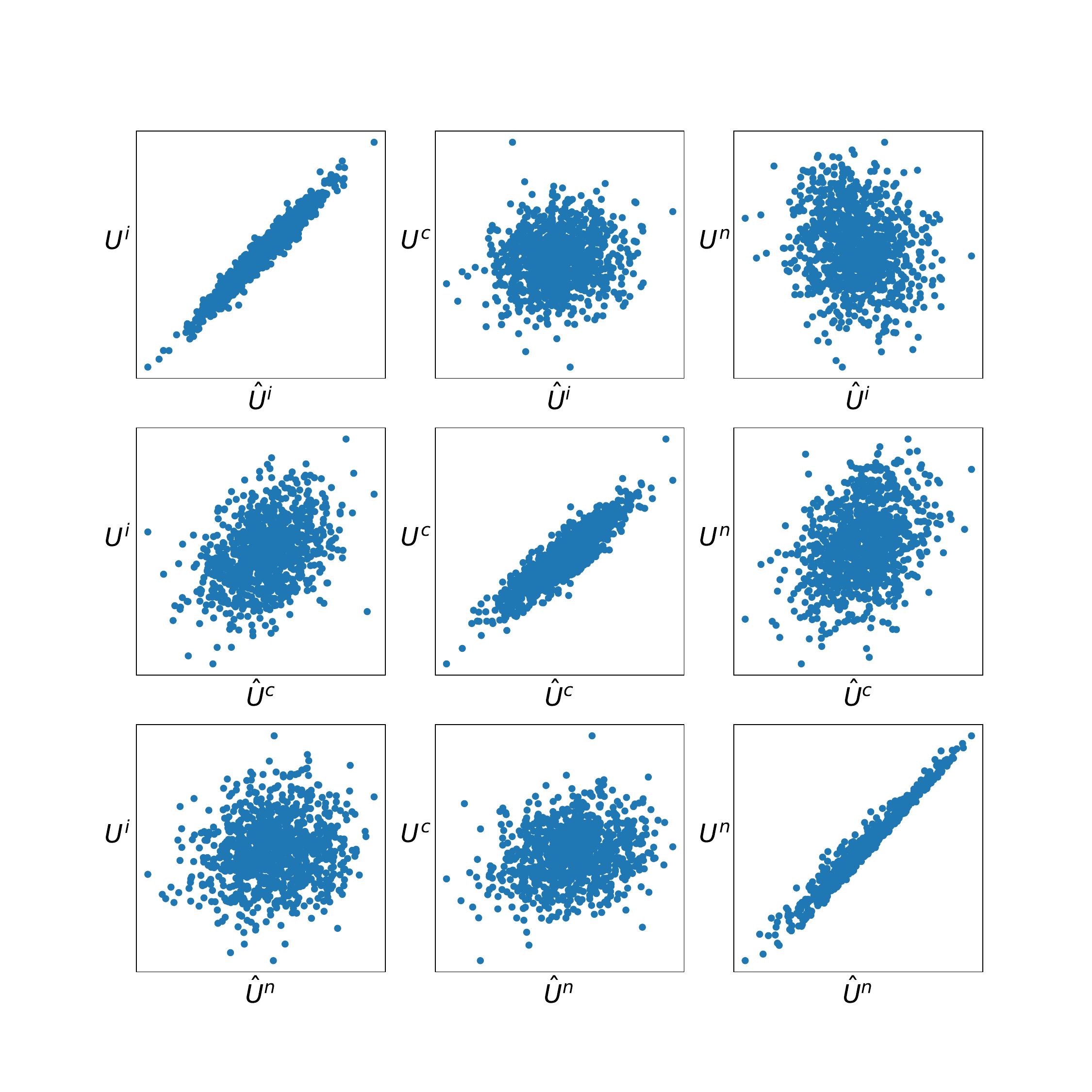}
    \caption{
    The figure presents scatter plots visualizing the relationship between the recovered latent confounders and ground-truth latent confounders $U^i, U^c$, and $U^n$.}
    \label{fig: recovered latent}
\end{figure}

\begin{table}[!h] 
\centering
\caption{MCC results between the recovered latent confounders and ground-truth latent confounders $U^i, U^c$, and $U^n$.}
\resizebox{0.35\textwidth}{!}{ 
    \begin{tabular}{cccc} 
        \hline
         &  $\hat U^{i}$ & $\hat U^{c}$ &$\hat U^{n}$ \\
        \hline
        ${U}^{i}$ &\textbf{ 0.9729} & 0.1742 & 0.2032 \\
        ${U}^{c}$ & 0.3796 & \textbf{0.9090} & 0.3616 \\
        ${U}^{n}$ & 0.1879 & 0.2420 & \textbf{0.9901} \\
        \hline
    \end{tabular}
}
\label{tab: mcc}
\end{table}

\subsubsection{\textbf{RQ2:  Correctness of Representation Learning}}

To validate the correctness of our representation learning method, we conduct experiments on the simulated dataset and visualize the recovered latent confounders $\hat U^i, \hat U^c, \hat U^n$ alongside ground-truth ones $ U^i,  U^c, U^n$ in Figure \ref{fig: recovered latent}.
We also report the corresponding Matthews Correlation Coefficient (MCC) results in Table \ref{tab: mcc}.
As shown in Figure \ref{fig: recovered latent}, the diagonal scatter plots (top-left, middle, and bottom-right) show strong alignments, indicating that each recovered latent confounder ($\hat{U}^i, \hat{U}^c, \hat{U}^n$) closely corresponds to its ground-truth counterpart ($U^i, U^c, U^n$).
In contrast, the off-diagonal scatter plots exhibit no obvious correlation, suggesting that the latent factors are well disentangled.
Similar findings are reflected in the MCC results in Table \ref{tab: mcc}, where the corresponding recovered and ground-truth confounders show high correlation, while the non-corresponding ones do not.
These results confirm that the proposed representation learning method effectively recovers the latent confounders, in line with the theoretical identifiability guarantees.

\begin{figure}[!h]
    \centering
    \includegraphics[width=0.32\linewidth]{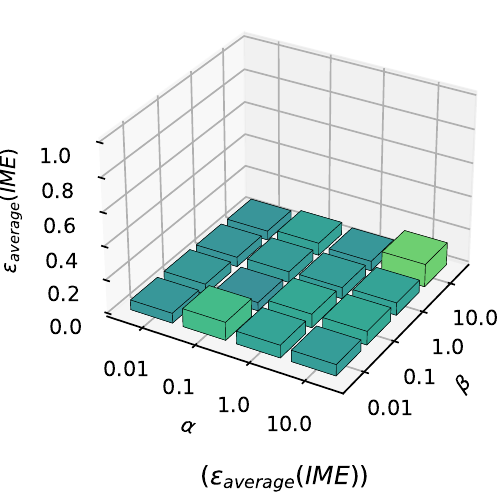}
    \includegraphics[width=0.32\linewidth]{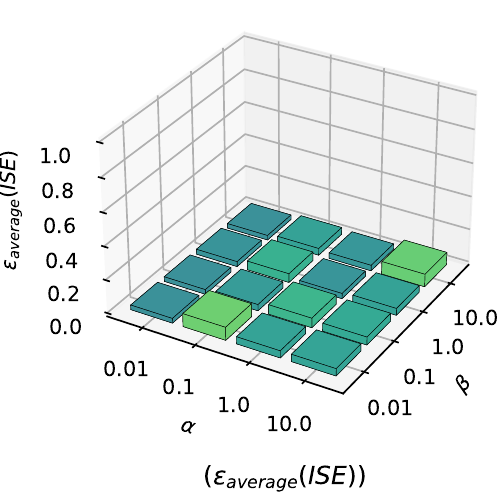}
    \includegraphics[width=0.32\linewidth]{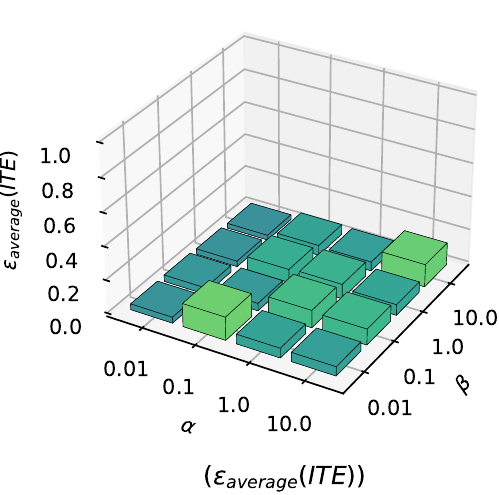}
    \caption{Hyperparameter sensitivity result regarding individual effect estimation error on BC(homo) dataset.}
    \label{fig: BC param ite}
\end{figure}

\begin{figure}[!h]
    \centering
    \includegraphics[width=0.32\linewidth]{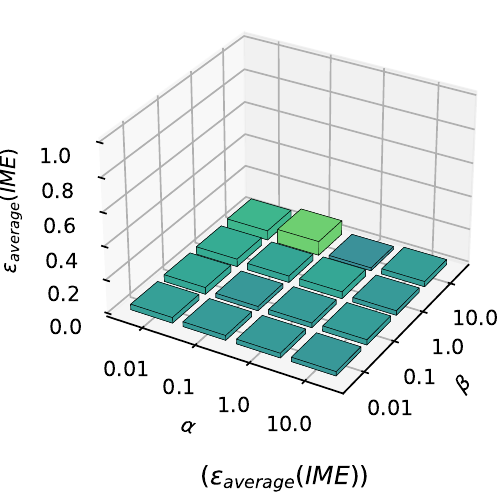}
    \includegraphics[width=0.32\linewidth]{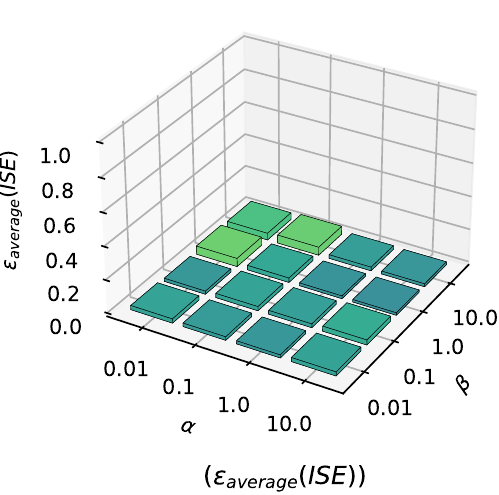}
    \includegraphics[width=0.32\linewidth]{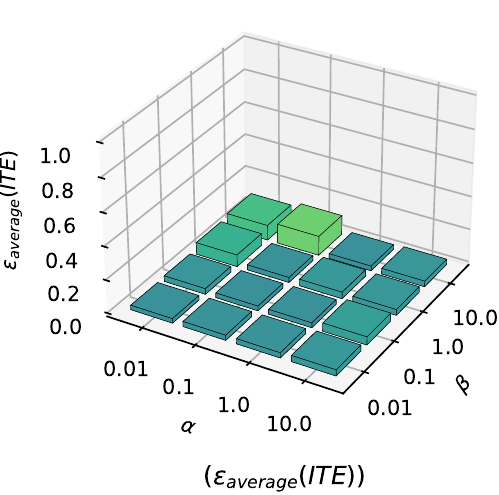}
    \caption{Hyperparameter sensitivity result regarding individual effect estimation error on Flickr(homo) dataset.}
    \label{fig: Flickr param ite}
\end{figure}

\subsubsection{\textbf{RQ3: Stability Regarding Hyperparameters}}
We evaluate the stability of CaLaNet under varying hyperparameter settings for $\alpha$ and $\beta$, choosing values from the set $\{0.01, 0.1, 1.0, 10.0\}$.
The results on individual effect estimation errors for both datasets are presented in Figures~\ref{fig: BC param ite} and~\ref{fig: Flickr param ite}, with comprehensive results available in Appendix E.
As illustrated in the figures, CaLaNet demonstrates relatively stable performance overall, suggesting that it is not highly sensitive to the choice of $\alpha$ and $\beta$ within a moderate range.
Also, we observe that excessively small or large values (e.g., $\alpha=0.01$ or $\beta=10.0$) can degrade performance, which is expected due to the resulting imbalance across different loss components.
In summary, CaLaNet performs well under reasonably balanced loss weights and shows reasonable stability across a wide range of hyperparameters.

\section{Conclusion}
\label{conclusion}

In this paper, we tackle the challenge of identifying and estimating causal effects under networked interference in the presence of latent confounders.
Specifically, we systematically categorize three distinct types of latent confounders that hinder networked effect identification.
To address this issue, we propose leveraging the networked information to achieve the identifiability of latent confounders.
Based on this recovery, we theoretically establish conditions under which networked causal effects become identifiable.
Building on these theoretical insights, we develop an effective estimator grounded in identifiable representation learning. 
Extensive experiments empirically validate both the theoretical results and the practical effectiveness of our proposed method.



\bibliographystyle{IEEEtran}
\bibliography{ieeetrans}



\begin{IEEEbiography}
[{\includegraphics[width=1in, height=1.25in, clip, keepaspectratio]{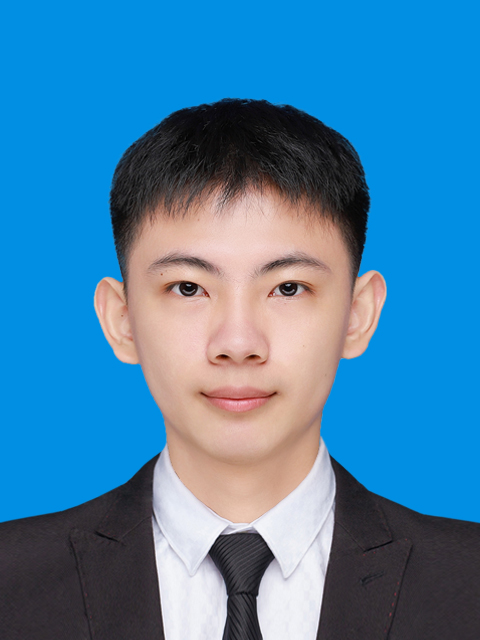}}]{Weilin Chen} is currently a postdoctoral researcher at the School of Intelligent Science and Technology, Peking University. He received his B.S. degree in Software Engineering in 2020 and his Ph.D. degree in Computer Science in 2025, both from Guangdong University of Technology, Guangzhou, China. In 2024, he was a visiting student at the University of Cambridge, UK. His current research interests include causal inference and its applications. He has served as a program committee member or reviewer for leading conferences and journals, including ICML, ICLR, NeurIPS, AISTATS, JMLR, and IEEE TNNLS, among others. 
\end{IEEEbiography}

\begin{IEEEbiography}[{\includegraphics[width=1in, height=1.25in, clip, keepaspectratio]{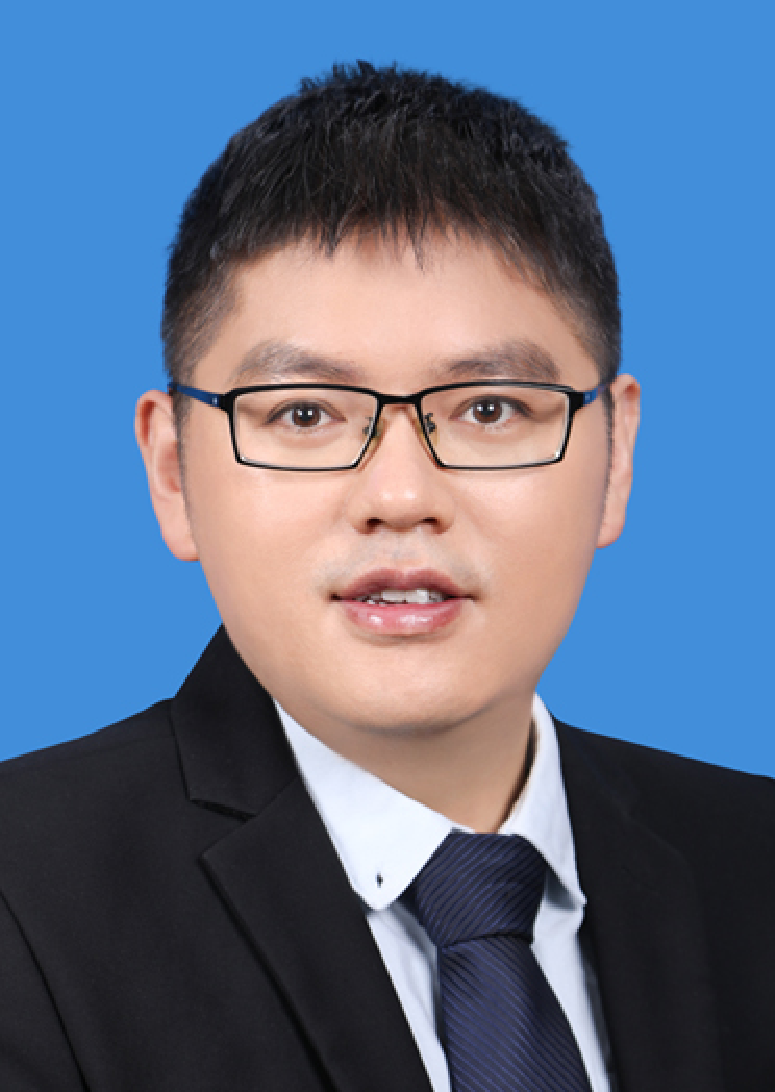}}]{Ruichu Cai} is currently a Full Professor in the School of Computer Science and the Director of the Data Mining and Information Retrieval Laboratory at Guangdong University of Technology. His research interests include causality, machine learning, foundation models, and their applications. He has received several awards, including the National Science Fund for Excellent Young Scholars and the Natural Science Award of Guangdong. He has served as an Associate Editor for IEEE TPAMI, an Action Editor for TMLR and Neural Networks, and an Area Chair for ICML, NeurIPS, and ICLR.

 
\end{IEEEbiography}

\begin{IEEEbiography}[{\includegraphics[width=1in,height=1in,clip,keepaspectratio]{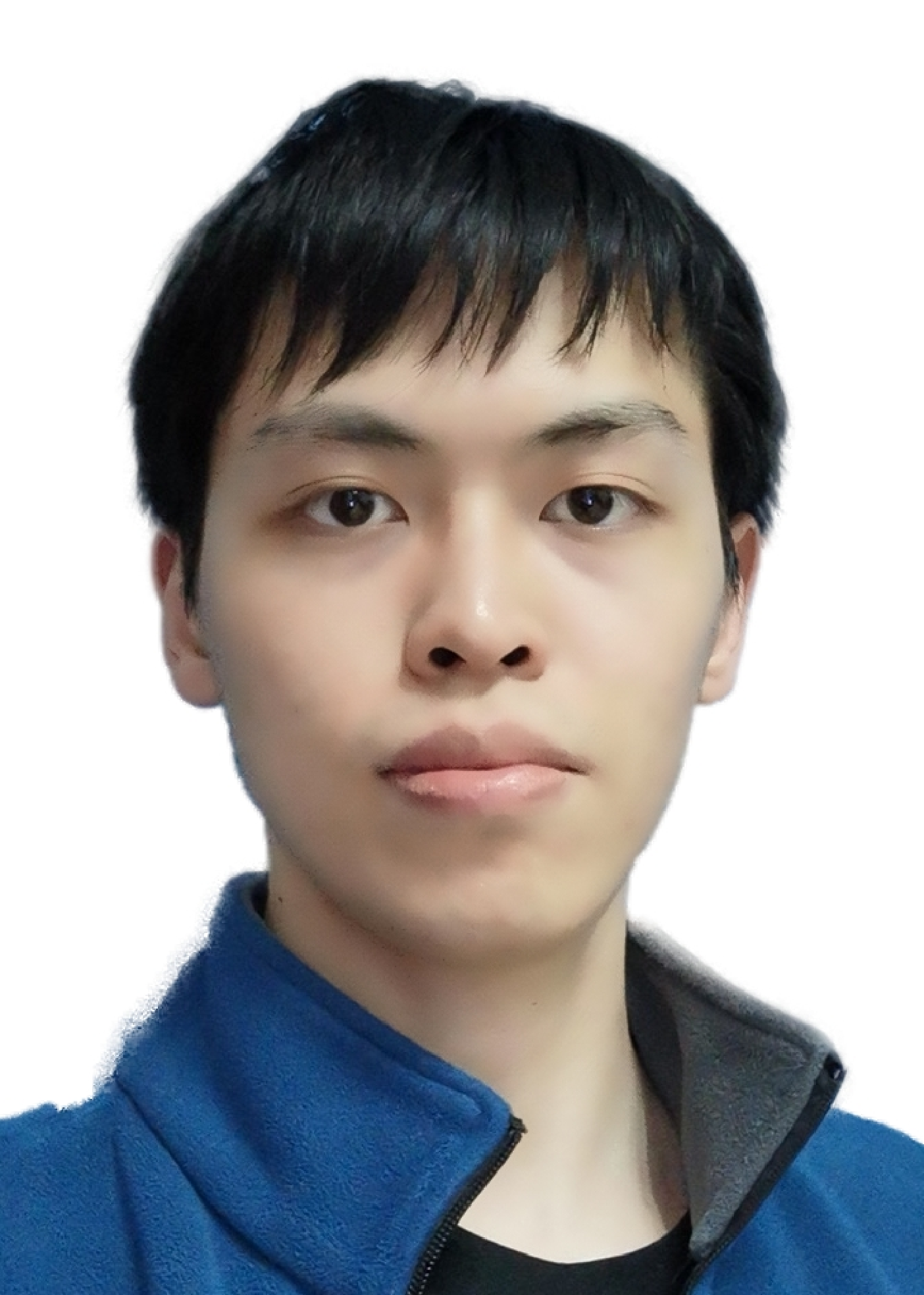}}]{Jie Qiao} received his Ph.D. from Guangdong University of Technology, School of Computer Science, in 2021. He is currently an Assistant Professor at the School of Computer Science, Guangdong University of Technology. His research interests cover various topics, including causality and its application. He has served as the program committee of ICML, NeurIPS, ICLR, AAAI, IJCAI, UAI, and so on.
\end{IEEEbiography}

\begin{IEEEbiography}[{\includegraphics[width=1in,height=1in,clip,keepaspectratio]{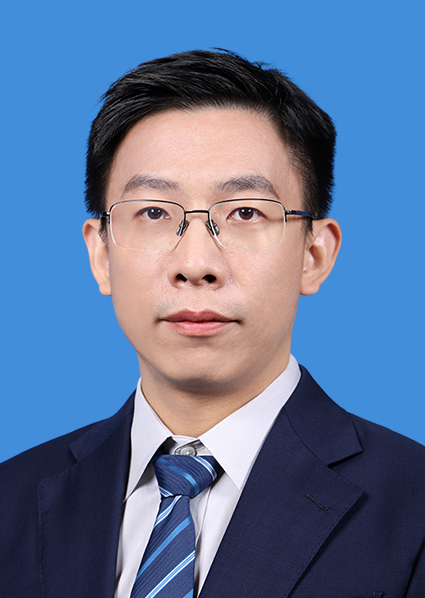}}]{Yuguang Yan} received the B.S. and Ph.D. degrees in the School of Software Engineering from South China University of Technology, China, in 2013 and 2019, respectively. He is currently with the School of Computer Science, Guangdong University of Technology, China.
He was a Post-doctoral Fellow at The University of Hong Kong from 2019 to 2021. His current research interests include optimal transport, causal inference, and graph data analysis.
\end{IEEEbiography}

\begin{IEEEbiography}
[{\includegraphics[width=1in, height=1.25in, clip, keepaspectratio]
{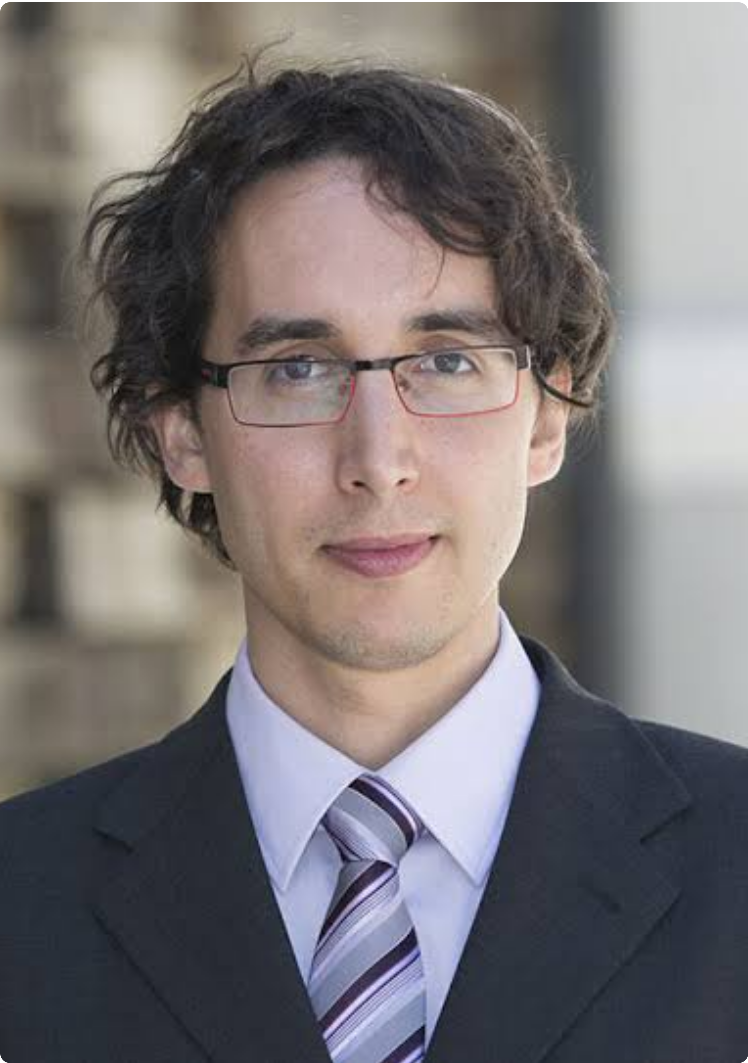}}]{Jos\'e Miguel Hern\'andez-Lobato} 
is Professor of Machine Learning at the Department of Engineering in the University of Cambridge, UK. Before joining Cambridge as faculty, he was a postdoctoral fellow at Harvard University, and before this, also a postdoctoral research associate at the University of Cambridge. Jose Miguel completed his Ph.D. and M.Phil. in Computer Science at Universidad Autónoma de Madrid (Spain), where he also obtained a B.Sc. in Computer Science from this institution, with a special prize to the best academic record on graduation. José Miguel's research interests are on probabilistic machine learning, with a focus on deep generative models, Bayesian optimization, approximate inference, causal inference, Bayesian neural networks and applications of these methods to real-world problems.

\end{IEEEbiography}

\vfill

\end{document}